\PassOptionsToPackage{small,bf}{caption}

\documentclass[smallextended]{svjour3}

\RequirePackage{fix-cm}
\usepackage{mathptmx} 
\smartqed
\journalname{Language Resources and Evaluation Journal}

\usepackage{caption}
\usepackage{subcaption}
\usepackage{url}
\usepackage{epsfig}
\usepackage{latexsym}
\usepackage{microtype}
\usepackage{tabularx}
\usepackage{amsmath}
\usepackage{arydshln}
\usepackage{booktabs}
\usepackage{wrapfig}
\usepackage{pgfplots}
\usepackage{textcomp}
\usepackage{chngpage}
\usepackage{graphicx,adjustbox}
\usepackage{multirow, makecell}
\usepackage[group-separator={,}, group-four-digits=true, group-digits=integer]{siunitx}
\usepackage[notext,not1]{stix}
\usepackage{fancyvrb}
\usepackage{natbib}
\usepackage{tabularx}
\usepackage{colortbl}
\usepackage{xcolor}
\usepackage[T5,T1]{fontenc}

\pgfplotsset{compat=1.7}

\usepackage{tikz}
\usepackage[linguistics]{forest}

\let\citewpar\citep
\let\citewopar\citet

\begin{document}

\title{DravidianMultiModality: A Dataset for Multi-modal Sentiment Analysis in Tamil and Malayalam 
}

\titlerunning{DravidianMultiModality}        

\author{Bharathi Raja Chakravarthi\(^1\)       \and
        Jishnu Parameswaran P.K\(^2\)\and
        Premjith B\(^2\) \and
        K.P Soman\(^2\) \and
        Rahul Ponnusamy\(^3\)\and
        Prasanna Kumar Kumaresan\(^3\)\and
        Kingston Pal Thamburaj\(^4\) \and
        John P. McCrae\(^1\) 
}


\authorrunning{Chakravarthi et al} 

\institute{Bharathi Raja Chakravarthi \\
              \email{bharathi.raja@insight-centre.org} \\          
         \\
         Jishnu Parameswaran P.K \\
         Premjith B \\
         K.P Soman \\
         Rahul Ponnusamy \\
         Prasanna Kumar Kumaresan \\
         Kingston Pal Thamburaj \\
         John P. McCrae    \\
          \(^1\)Insight SFI Research Centre for Data Analytics, Data Science Institute, National University of Ireland Galway, Galway, Ireland \\
          \(^2\)Center for Computational Engineering and Networking (CEN), Amrita School of Engineering, Coimbatore, Amrita Vishwa Vidyapeetham, India \\
           \(^3\)Indian Institute of Information Technology and Management-Kerala, India \\
            \(^4\)Universiti Pendidikan Sultan Idris, Malaysia
}

\date{Received: date / Accepted: date}

\maketitle

\begin{abstract}
Human communication is inherently multimodal and asynchronous. Analyzing human emotions and sentiment is an emerging field of artificial intelligence.  We are witnessing an increasing amount of multimodal content in local languages on social media about products and other topics. However, there are not many multimodal resources available for under-resourced Dravidian languages. Our study aims to create a multimodal sentiment analysis dataset for the under-resourced Tamil and Malayalam languages. First, we downloaded product or movies review videos from YouTube for Tamil and Malayalam. Next, we created captions for the videos with the help of annotators. Then we labelled the videos for sentiment, and verified the inter-annotator agreement using Fleiss's Kappa. This is the first multimodal sentiment analysis dataset for Tamil and Malayalam by volunteer annotators. This dataset is publicly available for future research in multimodal analysis in Tamil and Malayalam at Github\footnote{\url{githublink}} and Zenodo \footnote{\url{https://zenodo.org/}} .

\end{abstract}

\section{Introduction}
Computer Vision (CV) and Natural Language Processing (NLP), which offer computers the capability to grasp vision and language like humans, are recognised as necessary analysis fields of computer science \citewpar{JAIMES2007116}. As artificial intelligence becomes more integrated into everyday life throughout the world, intelligent beings capable of comprehending multimodal language in many cultures are in high demand.  Many multimodal analysis tasks have recently gained much attention since they bridge the boundary between vision and language to achieve human-level capability \citewpar{poria2017review}.  Complementary evidence from these unimodal characteristics is also appealing to multimodal content interpretation \citewpar{zadeh-etal-2017-tensor}. The abundance of behavioral clues is the fundamental benefit of studying videos over text analysis. In terms of voice and visual modalities, videos give multimodal data. Along with text data, voice modulations and facial expressions in visual data give key indicators to better detect real emotional states of the opinion bearer. As a result, combining text and video data aids in the development of a more accurate emotion and sentiment analysis model\citewpar{soleymani2017survey}. However, they are not developed for under-resourced languages.

Sentiment analysis in machine learning is typically modelled as a supervised classification task where the learning algorithm is trained on datasets that comprise sentiments such as positive, negative, or neutral in text \citewpar{wilson-etal-2005-recognizing,pak2010twitter,agarwal-etal-2011-sentiment,7122903}. However, human acquisition of sentiment does not occur only based on language but also based on other modalities \citewpar{multimodal-louis}.  Multimodal sentiment analysis aims to learn a joint representation of multiple modalities such as text, audio, an image or video to predict the sentiment of the content \citewpar{poria2017review}. This field aims to model real-world sentiment analysis from several distinct sensory inputs. Studies have shown that the use of vocabulary, vision and acoustic knowledge can boost the efficiency of Natural Language Processing (NLP) models.   The primary value of studying videos over textual content is the abundance of behavioural signs for identifying thoughts and feelings from different modalities. Although textual sentiment analysis facilities only use sentences, phrases and relationships, as well as their dependencies, they are deemed inadequate to derive related full-fledged sentiment from textual opinions \citewpar{perez-rosas-etal-2013-utterance}. 

In the last decade, sentiment analysis has attracted the attention of many scholars due to the increasing amount of user-generated content on social media \citewpar{yue2019survey}. Understanding users' preferences, habits, and contents by sentiment analysis and opinion mining opened up opportunities for many companies. Sentiments aid in decision-making, communication, and situation analysis \citewpar{taylor-keselj-2020-e}. Recently, due to the popularity and availability of handheld devices with cameras, more videos express their sentiment and opinion about a product, people and places than text \citewpar{poria-etal-2017-context,gella-etal-2018-dataset}. For example, users create videos by recording their opinions of products using mobile cameras or webcam and post them on social media such as YouTube, TikTok and Facebook \citewpar{castro-etal-2019-towards,qiyang2019learning}.  Therefore, analysis of multimodal content becomes vital to understand the opinions. 

However, the majority of the work on sentiment analysis and opinion mining focused on a single modality such as text or audio. Although some progress has been made for multimodal sentiment analysis in the English language \citewpar{multimodalporia,poria-etal-2019-meld}, this area still appears to be at the very early stages of research for other under-resourced languages including Dravidian languages. In this article, we are particularly interested in the Dravidian languages because of the importance and population of the native speaker of these languages.  Our work is the first attempt to create a multimodal video dataset for Dravidian languages to the best of our knowledge. Furthermore, most of the existing corpora for the Dravidian language are not readily available for research. 

This paper contributes to introducing a multimodal dataset of sentiment analysis for Tamil and Malayalam. Our dataset contains 134 videos, out of which 70 are Malayalam videos and 64 are Tamil videos. Each video segment includes manual transcription aligned along with sentiment annotation by volunteer annotators. The paper proceeds as follows: in Section \ref{relatedwork} we give a brief overview of related work in multimodal sentiment analysis datasets; in Section \ref{dravidian} we provide more in-depth background to Dravidian languages; in Section \ref{datacollection} the process for the corpus collection is described, including the challenges and methods used to overcome it; in Section \ref{Postprocessing} we explain the postprocessing methods we used; in Section \ref{Sentiment annotation} we outline the annotation of the corpus; in Section \ref{Corpusstat} we briefly describe corpus statistics, and in Section \ref{Conclusion} we conclude the paper and provide directions for future work.

\section{Related Work} \label{relatedwork}
Sentiment analysis and opinion mining have become an immense opportunity to understand users' habits and contents. Multimodal sentiment analysis, which combines verbal and nonverbal behaviours, has become one of the favourite research subjects in natural language processing \citewpar{baltruvsaitis2018multimodal,challenge-hml-2020-grand}.  The abundance of behavioural cues is the primary benefit of multimodal analysis videos over just uni-modal text analysis for detecting feelings and sentiment from opinions.  Many studies have focused on developing a new fusion network based on this structure to capture multimodal representation further \citewpar{cambria2017benchmarking,williams-etal-2018-dnn,blanchard-etal-2018-getting,sahay-etal-2020-low}. 

To study multimodal sentiment analysis and emotion recognition,  datasets were created by many research groups; notable works in this area are presented below. Interactive Emotional Dyadic Motion Capture Database (IEMOCAP) \citewpar{busso2008iemocap}  was created to understand expressive human communication by joint analysis of speech and gestures. This dataset was collected by the Speech Analysis and Interpretation Laboratory (SAIL) at the University of Southern California (USC). The dataset consists of 10K videos with sentiment and emotion labels recorded by ten actors. To do multimodal research in a real-world setting, a novel corpus for studying sentiment and subjectivity in opinion from Youtube Videos were created by \citewopar{morency2011} for the English language. This corpus was one of the earliest multimodal data created to show that it is feasible to benefit from tri-modal sentiment analysis, exploiting visual, audio, and textual modality. It was annotated for sentiment polarity by three workers for 47 videos and manual transcriptions for the audio data. 

\citewopar{rosas-spanish} created 105 Spanish videos annotated for sentiment polarity at utterance level.  CMU-MOSI \citewpar{zadeh2016mosi} dataset was created by CMU. It was annotated for sentiment in the range of [-3,3] for 2199 opinion video clips. AMMER \citewpar{cevher2019towards} is a German emotional recognition dataset about an in-car experiment that focuses on the drivers' interaction with both a virtual agent as well as a co-driver. CUM-MOSEI \citewpar{bagher-zadeh-etal-2018-multimodal} is a multimodal opinion sentiment and emotion intensity dataset that contains more than 1000 online YouTube speakers and more than 23,500 sentences from various topics and monologues. It also contains SDK to load the datasets into TensorFlow and PyTorch formats. This dataset was released for Grand Challenges \citewpar{ws-2018-grand}. This encouraged many researchers to do research on multimodal analysis. \citewopar{yu-etal-2020-ch} created a Chinese multimodal sentiment analysis dataset containing 2,281 refined video segments. For Spanish, Portuguese, German and French, \citewopar{bagher-zadeh-etal-2020-cmu} created a large-scare multimodal language dataset covering diverse set topics and speakers with 40,000 labelled sentences. 

While there are many datasets available for English and other languages, no multimodal video dataset annotated for sentiment analysis is available for the Tamil and Malayalam languages to the best of our knowledge.  We created a meme dataset for the Tamil language \citewpar{suryawanshi-etal-2020-dataset}; the dataset contains images along with text embedded on them, does not contain video.  We also  conducted a shared task to improve research on multimodal for Tamil \citewpar{suryawanshi-chakravarthi-2021-findings}. In this paper, we created a dataset contains 134 videos, out of which 70 are Malayalam videos, and 64 are Tamil videos. \citewopar{10.1109/MIS.2013.34} created a sentiment analysis of online videos containing movie reviews from YouTube. The videos are annotated at the video level for the sentiment. Our approach followed a similar approach and annotated sentiment at the video clip level.  Finally, we note that our dataset is not extensive as some of the previous dataset for other languages, since volunteers annotated it, but it has its own merit. We hope that this work will help to inspire other researchers to do work in multimodal analysis in Tamil and Malayalam languages. 

\section{Dravidian Languages} \label{dravidian}
Robert Caldwell was the first to use `Dravidian' as the common term for the prominent language family spoken in South India \citewpar{krishnamurti2003dravidian,steever2018tamil}. The new name was derived from the Sanskrit word Dravida or Dramila, which had previously been used to refer to the inhabitants of modern-day Tamil Nadu, Kerala, southern regions of Andhra Pradesh and Karnataka \citewpar{caldwell1875comparative,zvelebil1973smile}. For this paper, we will consider the present-day classification of languages in the Dravidian language family, including Brahui, Kannada, Kurukh, Telugu, Tulu, Malto, and more includes not only the south Indian language but also northern Indian, Sri Lankan and Pakistan languages\footnote{\url{https://en.wikipedia.org/wiki/Dravidian_languages}}. Dravidian languages are divided into four groups, South, South Central, Central, and North Dravidian. Major literary languages of the Dravidian family are Tamil, Malayalam, Kannada and Telugu, all belonging to the South and South Central group. We have created a resource for Tamil (ISO 639-3: tam) and Malayalam (ISO 639-3: mal), which belong to the South Dravidian group. 

The Tamil language is spoken in Tamil Nadu, India, Sri Lanka, Singapore, Malaysia, South Africa, and the Tamil diaspora \citewpar{sajeetha9063341,sajeetha9185369,sajeetha9342640}. The Tamil language is one of the official languages of Tamil Nadu, Puducherry, India, Sri Lanka, and Singapore comprises more than 80 million (2011-2019) speakers \footnote{\url{https://en.wikipedia.org/wiki/Tamil_language}}.  Malayalam is spoken by 33 million (2011-2019) \footnote{\url{https://en.wikipedia.org/wiki/Malayalam}} people in Kerala, Lakshadweep, Puducherry and other countries. The Tamil and Malayalam languages have their owns script, and they are different in many ways even though they are very closely related languages.  The Tamil script specified as Eluttu in most ancient grammar Tolkappiyam according to it 12  vowels (uyireluttu), 18 pure consonants (meyyeluttu) and one special character aytha eluttu, in total  31 letters in their independent form and an additional of 216 (12$\times$18) combinant letters, for a total of 247 (12+18+216+1) combinations (uyirmeyyeluttu) \citewpar{sakuntharaj2016novel,sakuntharaj2017use,sakuntharaj2018refined,sakuntharaj2018detecting}. 

Malayalam follows an abugida writing scheme \citewpar{mohanan1989syllable}. The writing system of the ancient Malayalam is known as Vatteluttu (round writing) because the characters are written circularly. The modern Malayalam script uses a transfigurated version of the Pallava Grantha script. Malayalam now has 15 vowels and 36 consonants; there are five pure consonant characters. Malayalam supports the generation of other character forms by combining consonants with vowels (14 $\times$36 - 3 = 501) \footnote{14 $\rightarrow$ excluding the $a$ sound, 3 $\rightarrow$\b{l}, \b{r} and \d{l} cannot be combined with \r{r}}, consonants with consonants (makes up around 57 characters \cite{krishnamurti2003dravidian}). Therefore, Malayalam has approximately 513 characters in it. The Pallava Grantha script was used to write Sanskrit in South India, and hence Malayalam has more letter than Tamil, which does not have voiced consonants and aspirates. 

Tamil is diglossic, which means its written form and spoken form are different \citewpar{schiffman1978diglossia,lokesh2019automatic}. The written form of present-day Tamil is standardized by the Government of Tamil Nadu, India and Sri Lanka\citewpar{schiffman1998standardization}.  However, there are many spoken forms, including Central Tamil dialect, Kongu Tamil, Chennai dialect, Madurai Tamil, Nellai Tamil, Kumari Tamil, Palakkad Tamil in India and Batticaloa Tamil dialect, Jaffna Tamil dialect, Negombo Tamil dialect in Sri Lanka \citewpar{zvelebil1960dialects,annamalai2015modern}. Kannada has heavily influenced the Sankethi Tamil dialect in Karnataka. Similarly, the Malayalam language also has many dialects. Malayalam dialects are mainly categorized based on geographical location and caste/religion. Dialects of Malayalam are: Malabar, Nagari-Malayalam, Malayalam, South Kerala, Central Kerala, North Kerala, Kayavar, Namboodiri, Mapilla, Pulaya, Nasrani, Nayar, and Kasargod. In addition to it, the people of Lakshadweep speak another dialect called Jeseri \citewpar{govindankutty1972proto}. We collected videos from social media YouTube where people from every region uploaded the video, so there many dialects in the study.

\section{Data collection} \label{datacollection}
This section presents the dataset prepared for the Multimodal Sentiment Analysis (MSA) in Dravidian languages such as  Malayalam and Tamil. All the video reviews were collected from YouTube.  The dataset comprises of the following: 
\begin{enumerate}
    \item Movie review videos in Malayalam and Tamil which have visual gestures required for multimodal sentiment analysis. 
    \item Speech transcriptions of each video for extracting the textual features for the analysis.
    \item Annotations for the data on a five-point scale - Highly Positive, Positive, Neutral, Negative and Highly Negative, which are the labels used for the classification. 
\end{enumerate}
The dataset contains 134 videos, out of which 70 are Malayalam videos and 64 are Tamil videos. We used various video downloading applications to download the videos from YouTube, and all of them have a fair resolution between 480 pixels and 720 pixels. 

\subsection{Acquisition of the videos for developing the dataset}

The dataset acquisition mainly focussed on movie reviews posted on YouTube by different vloggers. This work emphasised the development of the MSA dataset in Malayalam and Tamil, and hence we considered videos posted in those two languages only. However, there were no constraints imposed on the language in which a movie is made, which means that the dataset also contains Malayalam/Tamil reviews of other language movies. Furthermore, a few more conditions were fixed to choose the videos, which are listed below, 
\begin{itemize}
    \item Length of the video
    \\
    In this work, the length of all videos is fixed between one minute to three minutes. The videos with less than a minute length may not contain adequate visual features for identifying the sentiments. Similarly, lengthy videos may have more than one sentiment, which might make the task of sentiment analysis a difficult one. 
    \item The face of the reviewer
    \\Videos were collected in such a way that the face of a reviewer is clearly visible.  It helps to extract the required facial features from the data effectively. Therefore, videos with unclear faces and cropped faces were discarded from the selection of the dataset. 
    \item Background of the video
    \\
    In this work, we decided to select videos with a plain background because the textured background may affect the extraction of visual features from the text. 
\end{itemize}

The number of channels reviewing the movies in Malayalam and Tamil is less. In addition to that, these constraints set on selecting the preferred videos pose some challenges, which are listed below.

\begin{enumerate}
    \item Some reviewers review movies without showing their face but by displaying the movie posters. These videos cannot be accepted for the dataset because they do not contain any facial expressions. 
    
    \item Some videos contain both reviewer's face and the movie poster. Reviewers use movie posters or images to explain some part of the story or events in a movie from time to time. Because of this reason, it becomes difficult to capture a video of length at least one minute, which do not contain anything other than the reviewers face. 
    
    \item Some reviews contain more than one reviewer. There are two such scenarios.  
    
    \begin{itemize}
        \item In one scenario, two or more persons will always be present in the video. These two persons may have two different facial expressions depending on their views towards a particular movie. Here, the machine has to choose one of the faces and make predictions based on it.  How to select one face from multiple faces in a video is a question to be answered carefully. 
        
        \item In another scenario, two or more persons alternatively appears in the video, and the screen time each reviewer gets continuously is less than one minute. As a result of this, we were forced to exclude those videos from consideration.
    \end{itemize}
    
    \item Another type of reviews is the public review, where the reviewer asks the public about their opinion regarding the movies. We didn't consider those reviews for the stud, even though the number of such videos is high on YouTube. 
    
    \item Some videos satisfy all of our requirements but the background image. The videos with a plain background or minimum texture or graphics were considered because more textured background or graphics may prevent the algorithm from separating the face from the background.  
    
     \item Movie reviews shot outdoors also causes some issues with the background. In addition to the background, which contains various objects and colour schemes, such videos may comprise natural sounds. Such videos were also rejected from the selection. 
    
    \item Some reviewers do the vlogging while travelling, which causes a lot of background noise in the video. It also causes a shake in the video. These reasons prevented us from selecting such videos for preparing the dataset.   
    
    \item In some of the videos, the reviewers use more English words than Tamil or Malayalam words. The presence of more foreign language words affects the perception of speech sounds in Dravidian languages. Those videos were also not taken into account for dataset development.  
    
    \item A few videos were turned down because of the improper usage of microphones. It affected the audio quality of the review, and hence it became difficult to filter out the sound waves from the video.  
    
    \item Some reviewers use background music in their videos. It also caused difficulty in extracting the clean sound waves from the video. Since MSA also includes the analysis of speech signal, it is mandatory to have clean signals for extracting features for further analysis.  
\end{enumerate}

Finally, the videos which are devoid of all the problems mentioned above were considered for developing the dataset. Because of the underlying distribution complexity, diverse training samples are essential for thorough multimodal language investigations. The variety of intra-modal and cross-modal dynamics for language, vision, and auditory modalities lies at the basis of this complexity.  To ensure diversity, we considered videos spreading across all these categories regarding channels, age and gender.

\subsection{How did we overcome the challenges?}
This section explains the challenges we faced while collecting the data in different stages and the measures we took to tackle those issues. 

\subsubsection{Selection of the videos }

The first challenge comes from the selection of the videos. There were only a few vloggers doing videos of the type that we could accept. Using the limited channels available, we selected the videos with utmost care. After identifying the videos, the next challenge came from the quality of the videos. Some videos didn't have enough quality, which hindered the proper recognition of facial expressions. Those videos were downloaded with the highest possible quality and further scrutinised for selection. The sound quality of the videos was another hindrance to be solved. Some reviewers didn't use microphones, which resulted in below-par sound quality.  Therefore, videos with above-average quality audio were considered for the selection. Another challenge we faced was the presence of background noise. Videos that contain background noise that affect the audio quality were discarded. The next challenge comes from the volatile and high-tempo nature of the opinion videos, where speakers will often switch between topics and opinions. While they talk about an actor, they may suddenly change to music or cinematography. This makes it challenging to identify and segment the different opinions expressed by the speaker. The range, subtlety or intensity of sentiments expressed in the opinion videos was another obstacle to be faced.

In some cases, the way the reviewer verbalised his/her opinion did not match the objective of this work. Those situations come when the reviewer starts the video with negative remarks and concludes with a positive note without giving sufficient positive reasoning. Continuity in the explanation was another problem we found in the reviews. Some reviewers often change topics and deviate from the point they were discussing.

\subsubsection{Background of the video} 
Utmost care should be given to verify the background of the video. Reviewers may display movie posters on the screen for less than a second time, and it may not get noticed if we don't pay complete attention to the video. Some videos contain poster of movies or the channel logo, which also have to be removed. Further, instruments such as laptops, cameras, and microphones may also be there in the video. It is good to avoid such objects from a video to make it clean. Since the number of channels reviewing movies in Malayalam and Tamil is limited and the videos which satisfy all our constraints are much fewer, we cannot simply drop all the videos. Therefore, to make the videos with above-stated background issues acceptable for the dataset, we decided to crop the portions which contain movie posters, logo, and instruments.

\subsubsection{Editing the video} 

In most of the videos, the length ranges from 3 minutes to 15 minutes. Reviewers may talk about paid promotions, the story of the movie, and other things. But we are interested in the portion where the reviewer talks positively or negatively about the video. It is essential to locate the beginning and end of the sentence to do that. It brings another challenge to the length of the video. We decided to keep the minimum and maximum duration of the video to 1 minute and 3 minutes, respectively, for the dataset. Therefore, finding a video with relevant content whose length falls within the stipulated range is a strenuous task. In addition to that, reviewers sometimes stop mid-sentence and immediately start the following sentence. It was challenging to edit such portions because we have to find out the exact location (or time in minutes and seconds) where we have to cut. These processes are time-consuming as we have to spend time watching the videos several times to edit the desired portion.

\subsubsection{Preparation of the transcripts}

Preparation of the transcript was the most challenging task as it could only be done with human assistance. Initially, we used Google speech-to-text\footnote{https://cloud.google.com/speech-to-text/} and IBM speech-to-text \footnote{https://www.ibm.com/cloud/watson-speech-to-text} models for transcribing the videos. However, the results were not satisfactory due to the dialect, pronunciation, clarity in speech, and improper sentence construction. Therefore, we decided to prepare the transcripts manually. Following are the difficulties we faced while preparing the transcripts,
\begin{enumerate}
    \item Reviewers may speak fast, which makes it difficult to perceive.
    \item Reviewers stop sentences abruptly.
    \item Obscured pronunciation due to various reasons such as a slip of the tongue, local slang, and ignorance of the proper pronunciation. 
    \item To transcribe a video most efficiently, we had to watch it repeatedly and listen to the words one by one. Writing transcriptions in Malayalam and Tamil using a QWERTY keyboard is a laborious job because of the difficulty of identifying the keyboard characters. Therefore, we utilised Google input tools \footnote{https://www.google.com/inputtools} to prepare the transcripts. The Malayalam and Tamil transcripts written using Google input tools were copied into a Notepad and saved to the local drive. Figures \ref{fig:transcription} illustrates this process. 
    
    \begin{figure}
     \begin{subfigure}[b]{0.3\textwidth}
         \includegraphics[height= 5cm, width=\textwidth]{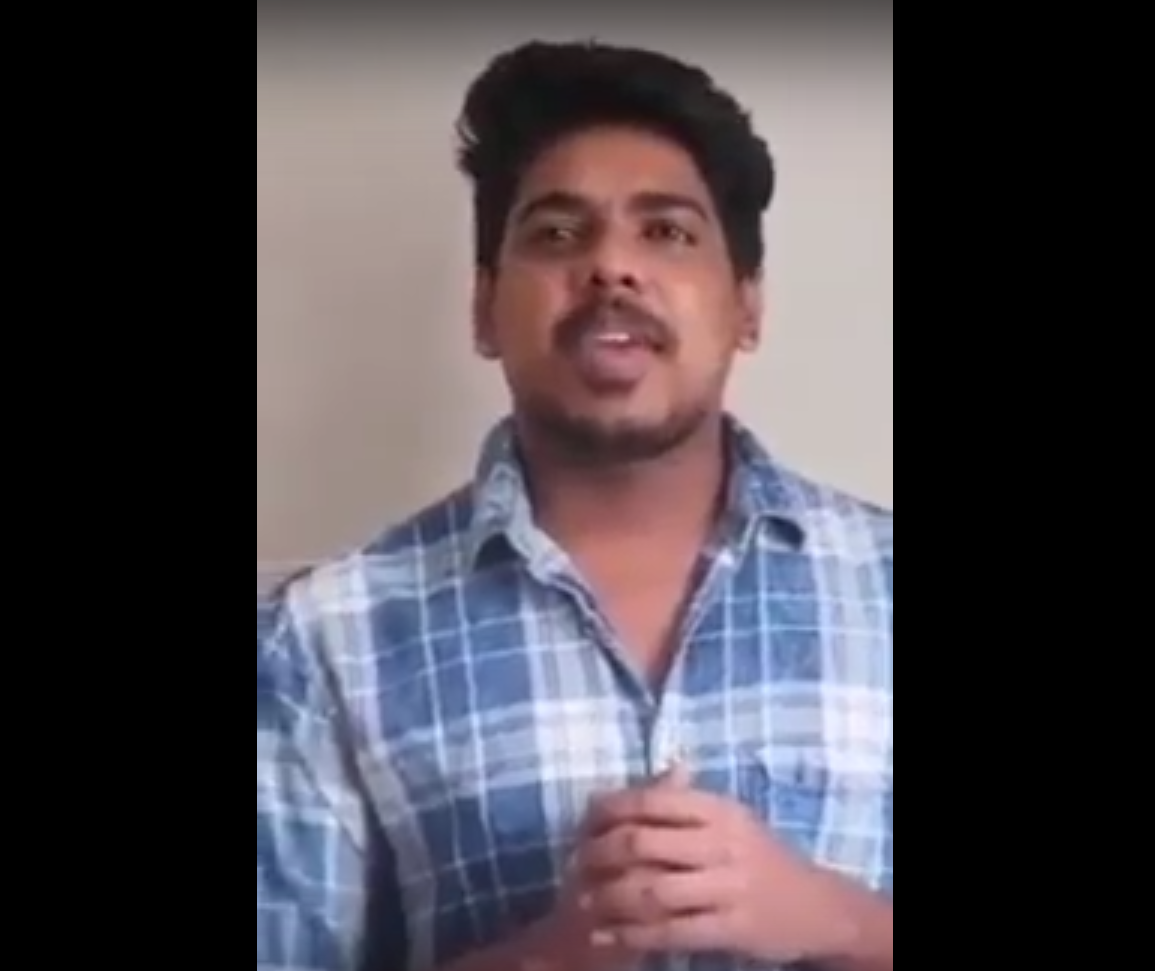}
     \end{subfigure}
     \begin{subfigure}[b]{0.3\textwidth}
         \includegraphics[height= 5cm, width=10cm]{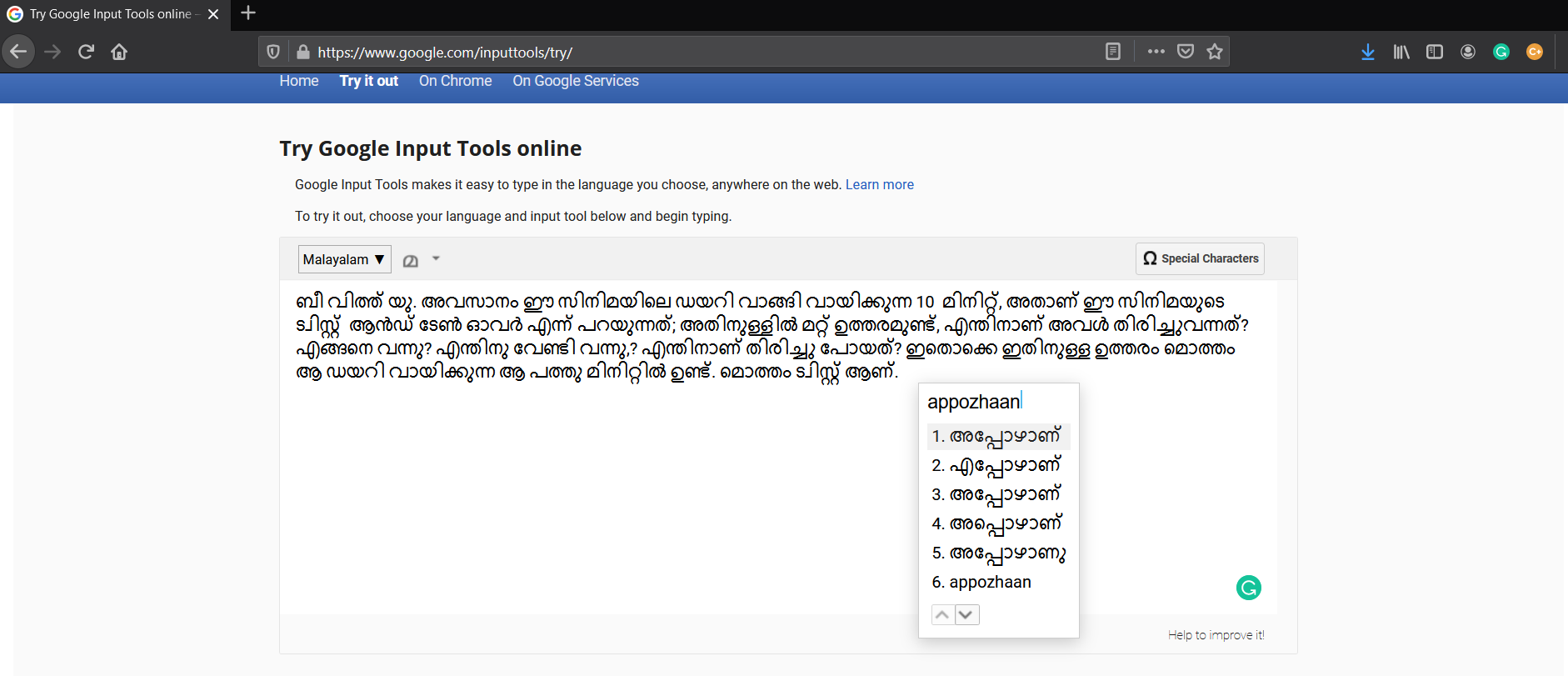}
     \end{subfigure}
        \caption{Screenshot of a video and preparing its transcription using Google input tools. }
        \label{fig:transcription}
\end{figure}

    
    
    \item Placing punctuation was another cumbersome task. Reviewers stop sentences without completing them and start the sentences without a proper beginning. In general, many sentences did not have a formal sentence structure. It leads to an ambiguous usage of punctuation marks in the transcripts. A decision on where to keep the punctuation and which punctuation to use was taken after watching the videos and listening to the audio several times.
    \item Identification of the dialects of the reviewers was another difficult job. There are diverse dialects in Malayalam and Tamil, and it is a demanding task for a person to understand those dialects if (s)he is not familiar with them. Therefore, we had to seek help from others to understand the words, and thereby the spelling. 
    \item Problems with the dialects affected the annotation process also. Most of the Malayalam and Tamil words possess diverse senses in different contexts. In some cases, the meanings are opposite also. Neologisms added complexity to this scenario. Nowadays, terms generally used for expressing negative senses are widely used for expressing positive feelings. It causes ambiguity while annotating the video if the annotator is unaware of such usages in a particular dialect. 
\end{enumerate}

The average time taken to complete the transcription for one video in Malayalam was 2 hours, whereas Tamil's case exceeds 3 hours, making the transcription time-consuming.

\section{Post-processing of the selected videos} \label{Postprocessing}

Post-processing is required to make the selected videos appropriate for the dataset for the  MSA in Dravidian languages. Following are the two post-processing steps we followed for preparing the dataset to match the criteria set for the dataset. 

\begin{figure}[h]
         \centering
         \includegraphics[width=\textwidth]{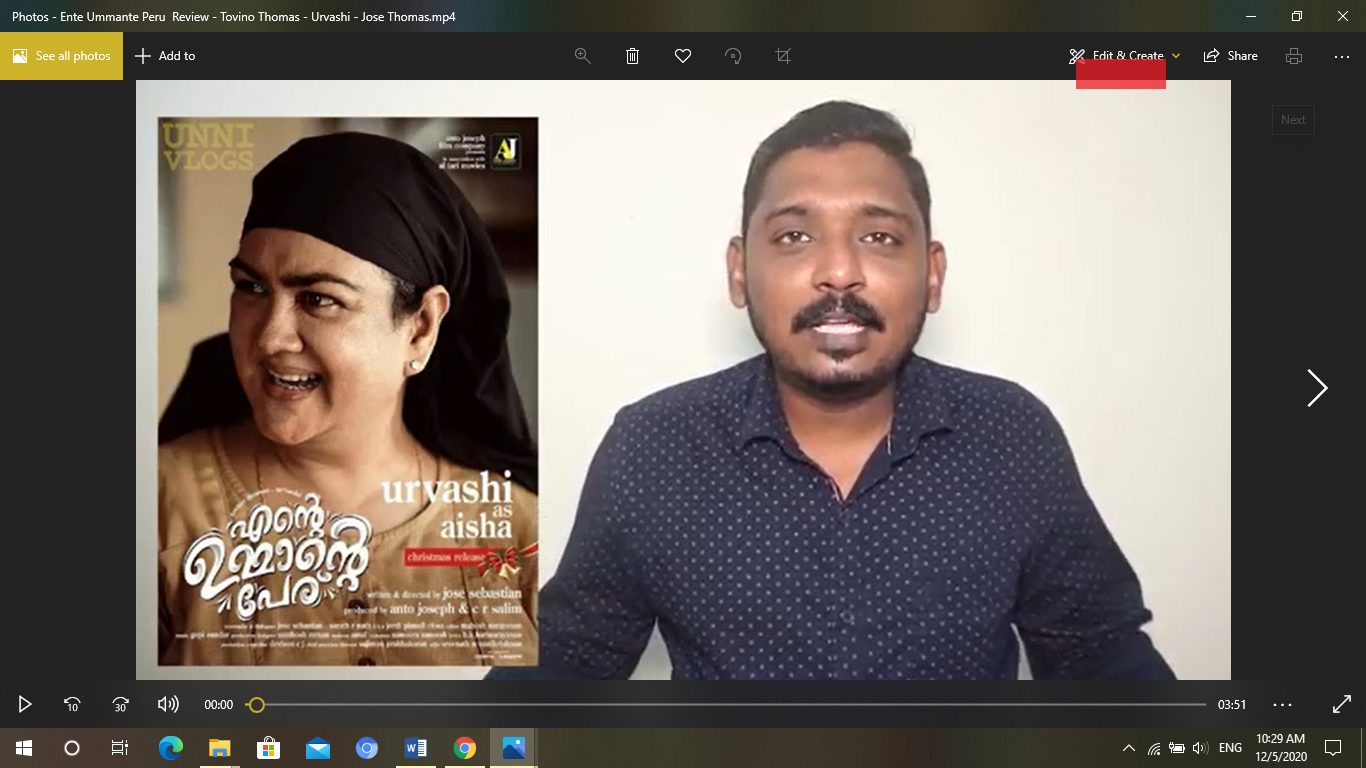}
         \caption{Choose Edit \& Create menu from the Windows Photos application}
         \label{fig:edit-create}
\end{figure}

\begin{figure}[h]
         \centering
         \includegraphics[width=\textwidth]{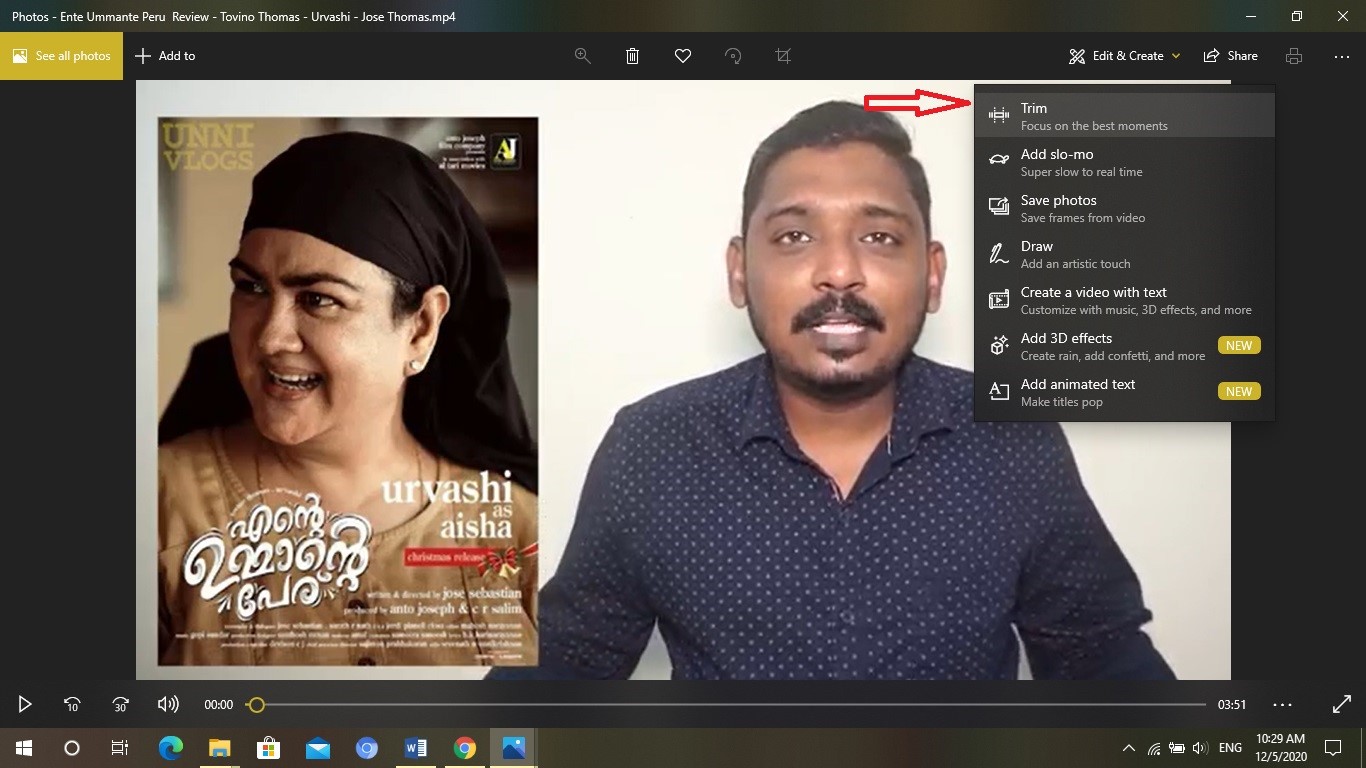}
         \caption{Choose the Trim option to cut the required portion from a video}
         \label{fig:trim}
\end{figure}

\begin{figure}[h]
         \centering
         \includegraphics[width=\textwidth]{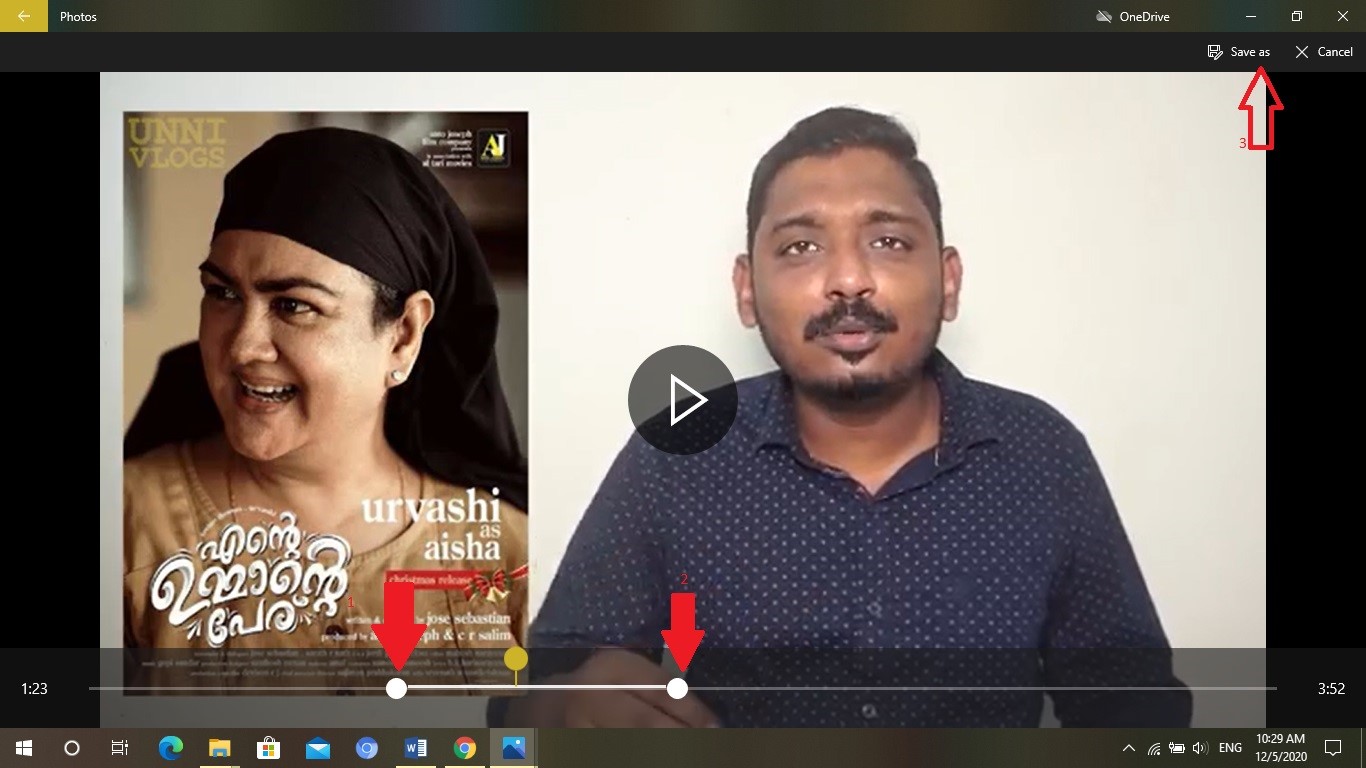}
         \caption{Fix the portion of the video to be edited out}
         \label{fig:fix-length}
\end{figure}

\subsection{Fixing the length of the video}

The length of the videos in the dataset has to be between 1 minute and 3 minutes. Therefore, the video duration had to be shortened to bring the length within the above-stated range. While doing so, it has to be made clear that the video starts with a proper sentence (should not start at the middle of a sentence) and ends with another. Moreover, the selected portion should convey an opinion about the movie. It has to be done by giving maximum attention. We used the Windows Photos application to edit the videos. 

Steps followed for editing the videos using the Photos application is given below,
\begin{enumerate}
    \item Open the video using the Photos application.
    \item Select the Edit \& Create option.
    \item Choose the Trim option from the drop-down menu and select the portion of the video to be extracted.
    \item Save the extracted video in the drive.
\end{enumerate}

Figures \ref{fig:edit-create}, \ref{fig:trim} and \ref{fig:fix-length} illustrates the steps we followed for editing the videos.

\subsection{Cropping of video}
\begin{figure}[htb]
\centering
\includegraphics[width=9cm, height=14cm]{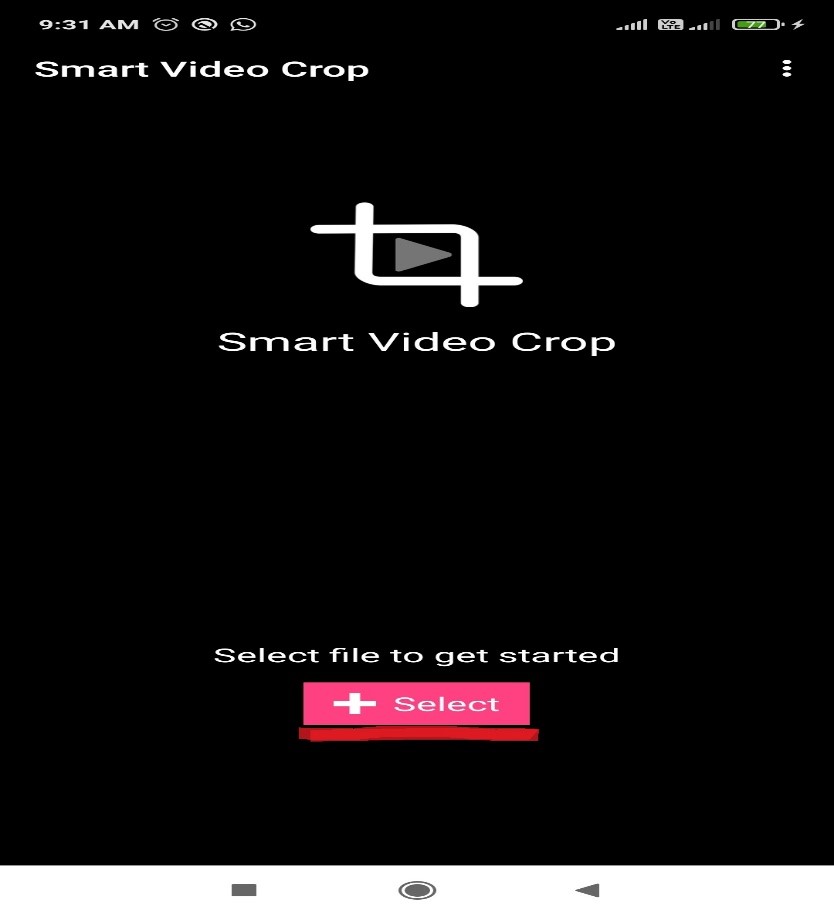} 
\caption{Selecting the video to be cropped.}
\label{fig:step-1}
\end{figure}

\begin{figure}[h]
\centering
\includegraphics[width=9cm, height=14cm]{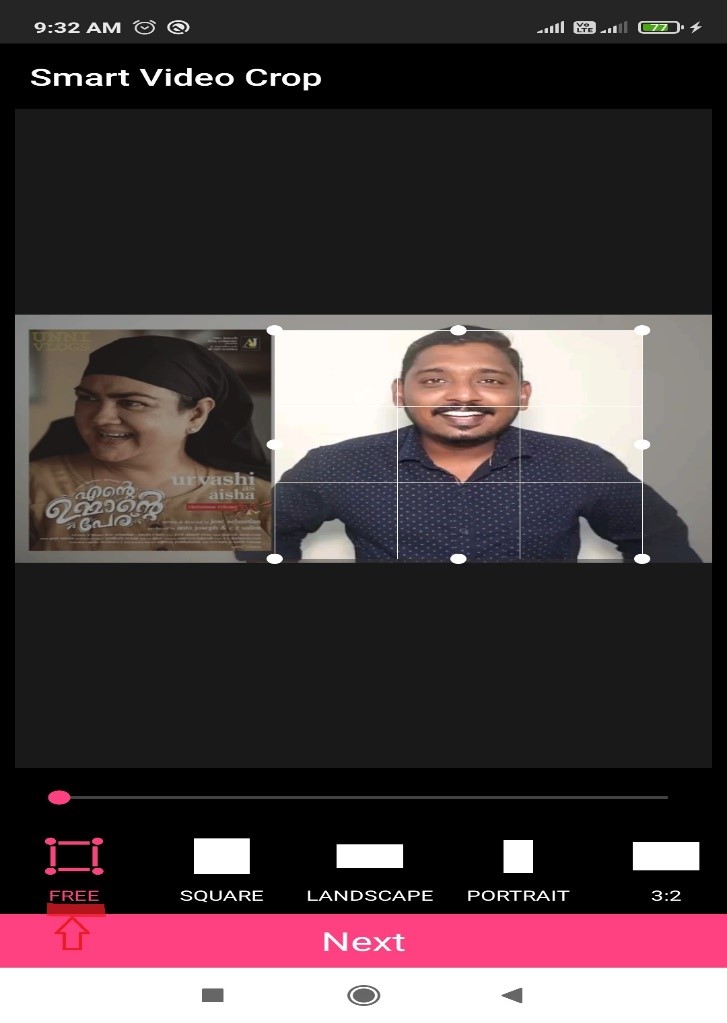}
\caption{Select the cropping option from the list of options available.}
\label{fig:step-2}
\end{figure}

\begin{figure}[h]
\centering
\includegraphics[width=9cm, height=14cm]{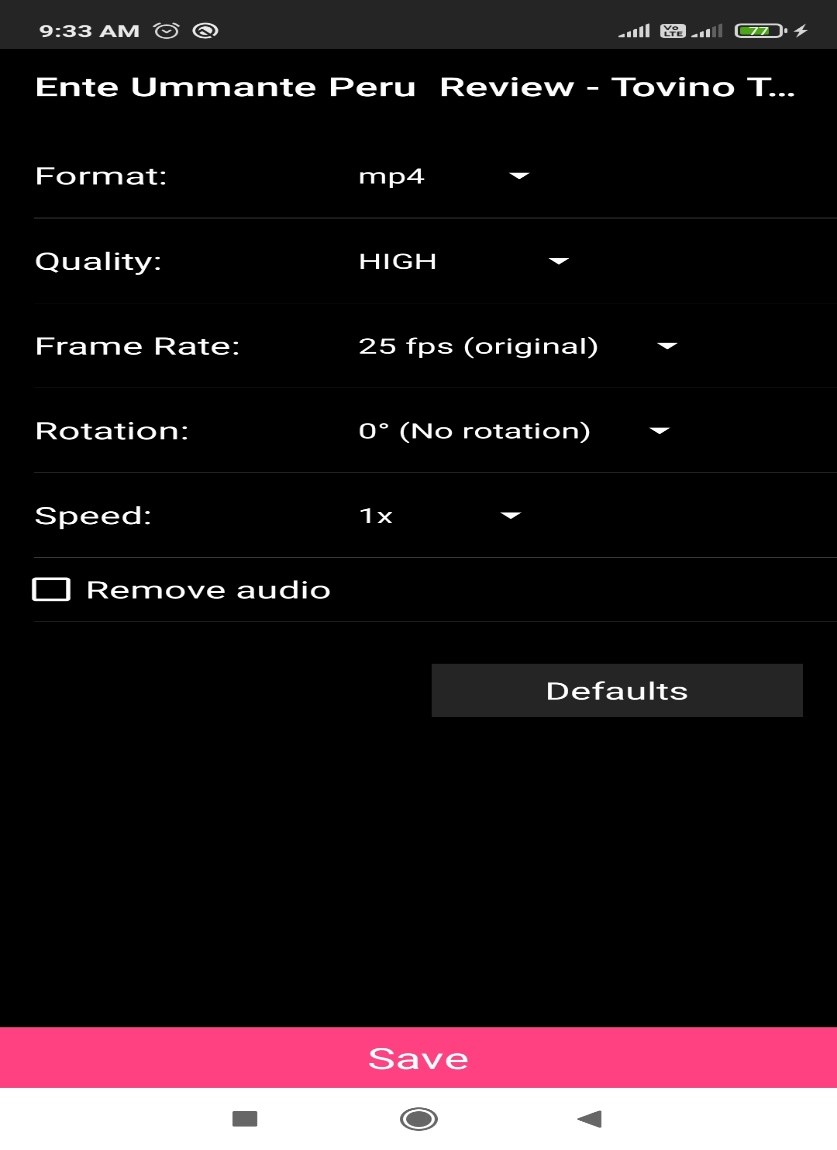}
\caption{Select the format and quality of the video to be saved}
\label{fig:step-3}
\end{figure}

\begin{figure}[h]
\centering
\includegraphics[width=9cm, height=14cm]{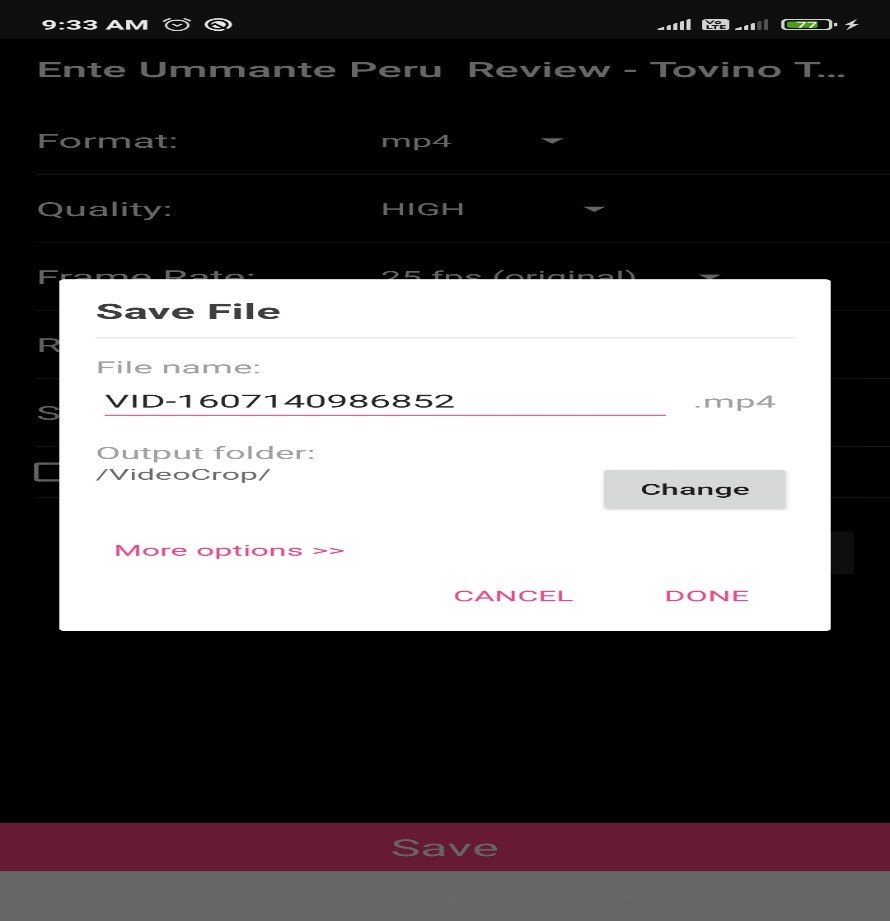}
\caption{Select the location in the drive to save the file}
\label{fig:step-4}
\end{figure}

\begin{figure}[h]
\centering
\includegraphics[scale=0.05]{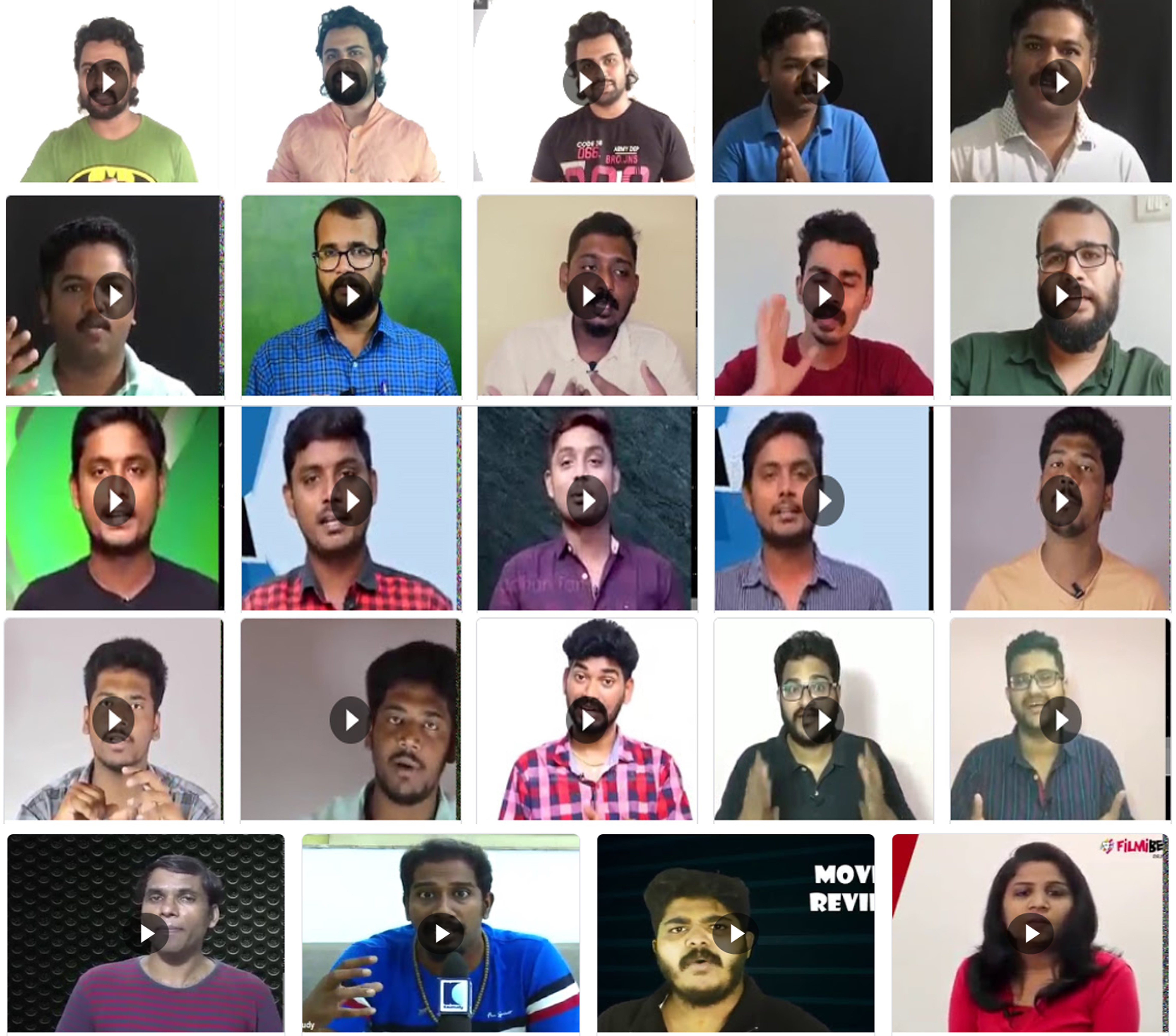}
\caption{Screenshots of the selected after the post-processing}
\label{fig:step-5}
\end{figure}

This post-processing step was carried out to remove the advertisements, movie posters, electronic gadgets and other properties, logos and ratings for the movie. We employed this process on the videos which are found to be appropriate for the dataset. It can be done with any video cropping application, and we used an Android application named Smart Video Crop\footnote{https://play.google.com/store/apps/details?id=com.clogica.videocrop\&hl=en\&gl=US} to crop out the unnecessary portions from the video. We checked the quality of the videos after this process to make sure that both video and audio quality are not degraded. Steps involved in cropping the videos using Smart Video Crop are illustrated in Figures \ref{fig:step-1}, \ref{fig:step-2}, \ref{fig:step-3} and \ref{fig:step-4}.   Screenshots of the selected and cropped videos are shown in Figure \ref{fig:step-5}.

\section{Sentiment Annotation} \label{Sentiment annotation}

The final step in the dataset development is annotation. It is also a complicated job as it requires the opinions of multiple annotators to finalise the labels for each data. In this work, we annotated the data with five sentiments - Positive, Highly Positive, Neutral, Negative, and Highly Negative. 

\subsection{Annotation Setup}
We took the procedures put forward by \cite{zadeh2016mosi} and \cite{mohammad-2016-practical} into consideration for annotating the video clips. Three annotators annotated each video clip for each language.  Unlike unimodal data, multimodal data needs a more detailed analysis of the data by considering the video, audio and text for annotation. The facial expression of the reviewer is substantial to determine the sentiment. We considered clues obtained from both facial expressions of the reviewer and his/her vocal modulations (and transcript) to annotate the video. Considering the facial expression, the tone of the speech helps to distinguish the sentiments of the video clips, especially when the reviewer uses sarcastic comments. The annotation schema is given below.
\begin{itemize}
    \item \textbf{Positive state:}  A video clip is annotated as positive if the reviewer uses positive words with mild facial expression to review that video. 
    \item \textbf{Highly positive state:} If the reviewer uses overstated words or expressions to describe a movie, we annotated it as a highly positive movie. 
    \item \textbf{Negative state:} There is a usage of negative words and sarcastic comments with soft facial expressions in a video that helped to label it as negative. 
    \item \textbf{Highly negative state:} Similar to the highly positive state, if the reviewer exaggeratedly uses negative words with a sullen face and taut voice, we consider those video clips as highly negative ones. 
    \item \textbf{Neutral state:} There is no explicit or implicit indicator of the speaker’s emotional state: Examples are asking for like or subscription or questions about the release date or movie dialogue. This state can be considered a neutral state.

\end{itemize}

\subsection{Annotators}

We shared separate Google sheets with each of the annotators to avoid copying the labels in case of any ambiguity in identifying the sentiment. The annotators include one female and two male for both languages. Except for one Tamil annotator, all other annotators were postgraduates. Malayalam annotators were proficient in Malayalam and English, whereas all the Tamil annotators were polyglots with competence in Tamil, Malayalam and English. Among the annotators, only two (in Tamil) did schooling in English medium and others in their native language.  Table \ref{tab:annotators} shows the details of the annotators. Annotators, except one, are students of Amrita Vishwa Vidyapeetham, India, who volunteered to do the annotation. One Tamil annotator is a parent of the author who can read, write and speak Tamil.  

\begin{table}[t] 
\begin{center} 
\begin{tabular}{|l|l|l|l|l|}
\hline
& & Malayalam & Tamil\\
\hline
Gender & Male & 2 & 2  \\
& Female & 1 & 1 \\
\hline
Highest Education & Undegraduate & 0 & 1 \\
& Postgraduate & 3 & 2 \\
\hline
Medium of Schooling & English & 0 & 2  \\
& Native & 3 & 1 \\
\hline
Total & & 3 & 3 &\\
\hline
\end{tabular}
\caption{Details of the annotators who did annotation of the video clips.} 
\label{tab:annotators} 
\end{center} 
\end{table}

Labelling was done manually by considering the following aspects after observing the video clips using VLC video player \footnote{https://www.videolan.org/vlc/}, 

\subsection{Facial expression and gestures}

\begin{figure}
     \centering
     \begin{subfigure}[b]{0.3\textwidth}
         \centering
         \includegraphics[height= 3.1cm, width=\textwidth]{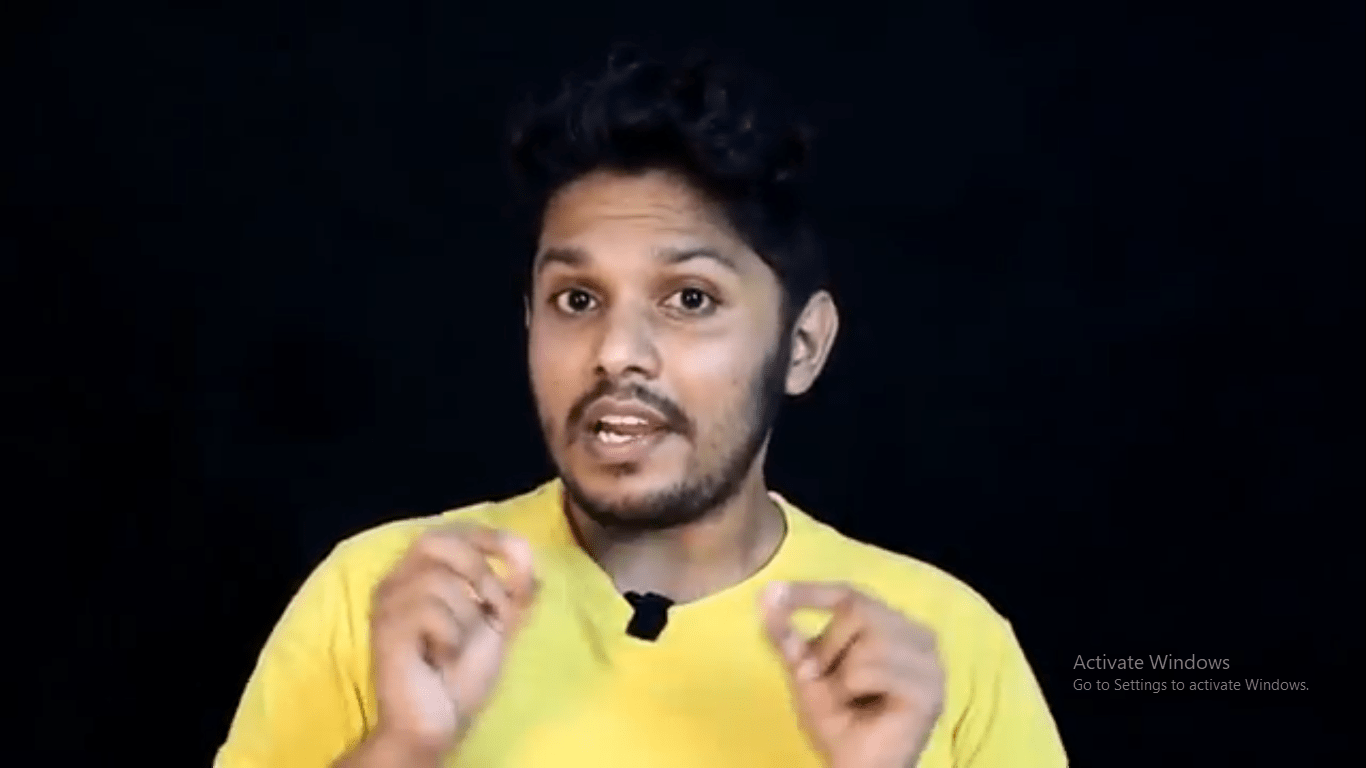}
     \end{subfigure}
     \begin{subfigure}[b]{0.3\textwidth}
         \centering
         \includegraphics[width=\textwidth]{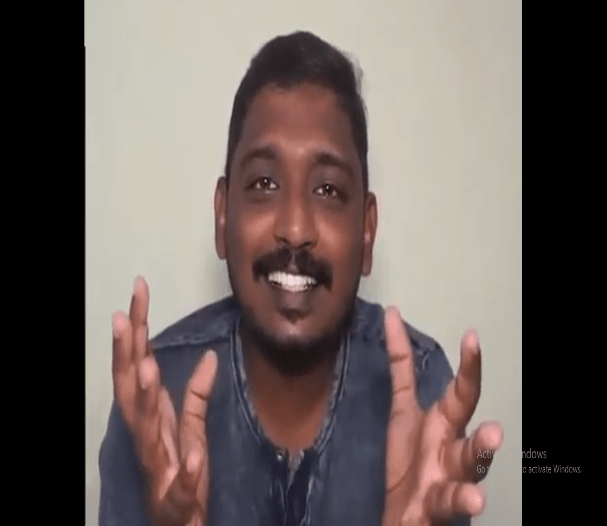}
     \end{subfigure}
        \caption{Screenshots of Malayalam video clips with highly positive sentiment}
        \label{fig:high-pos-mal}
\end{figure}

\begin{figure}
     \centering
     \begin{subfigure}[b]{0.3\textwidth}
         \centering
         \includegraphics[height= 3cm, width=\textwidth]{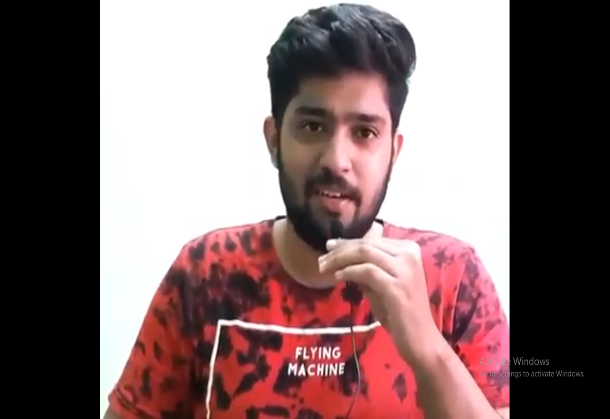}
     \end{subfigure}
     \begin{subfigure}[b]{0.3\textwidth}
         \centering
         \includegraphics[height= 3cm, width=\textwidth]{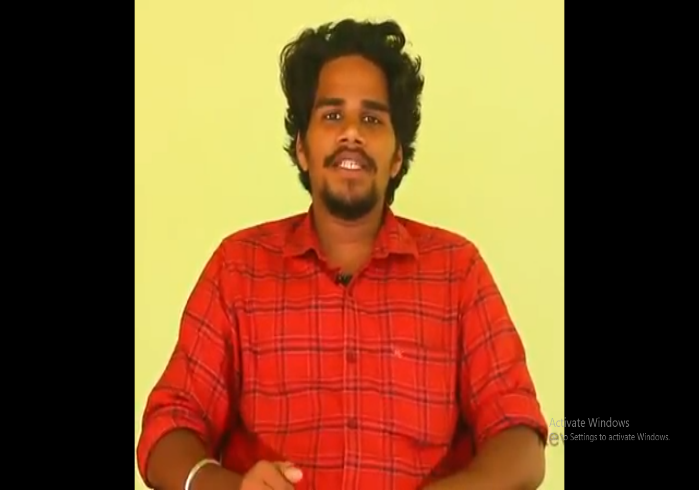}
     \end{subfigure}
        \caption{Screenshots of Malayalam video clips with positive sentiment}
        \label{fig:pos-mal}
\end{figure}

\begin{figure}
     \centering
     \begin{subfigure}[b]{0.3\textwidth}
         \centering
         \includegraphics[height= 3cm, width=\textwidth]{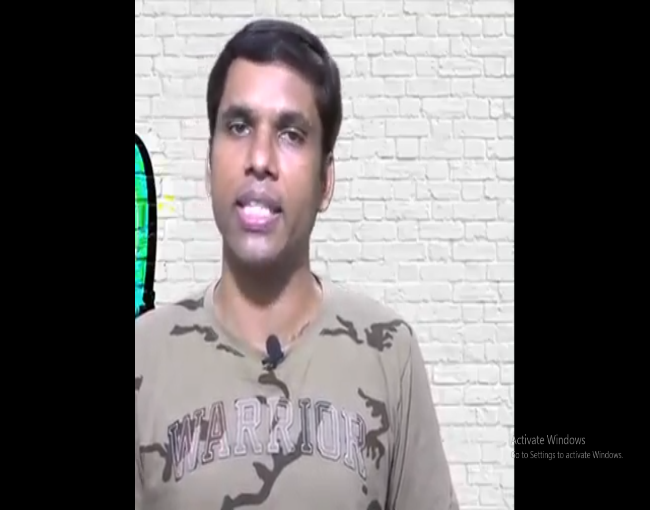}
     \end{subfigure}
     \begin{subfigure}[b]{0.3\textwidth}
         \centering
         \includegraphics[height= 3cm, width=\textwidth]{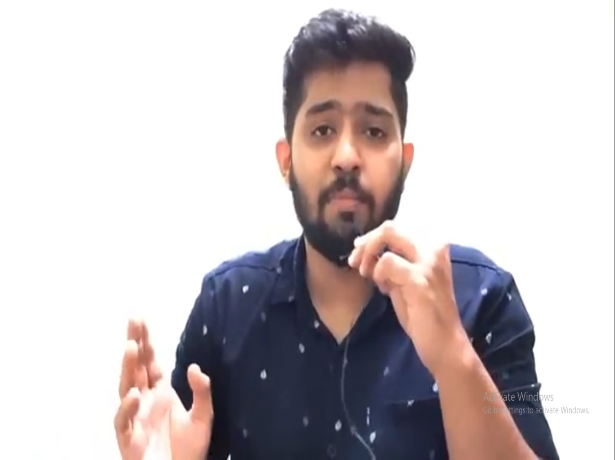}
     \end{subfigure}
        \caption{Screenshots of Malayalam video clips with neutral sentiment}
        \label{fig:neu-mal}
\end{figure}

\begin{figure}
     \centering
     \begin{subfigure}[b]{0.3\textwidth}
         \centering
         \includegraphics[height= 3cm, width=\textwidth]{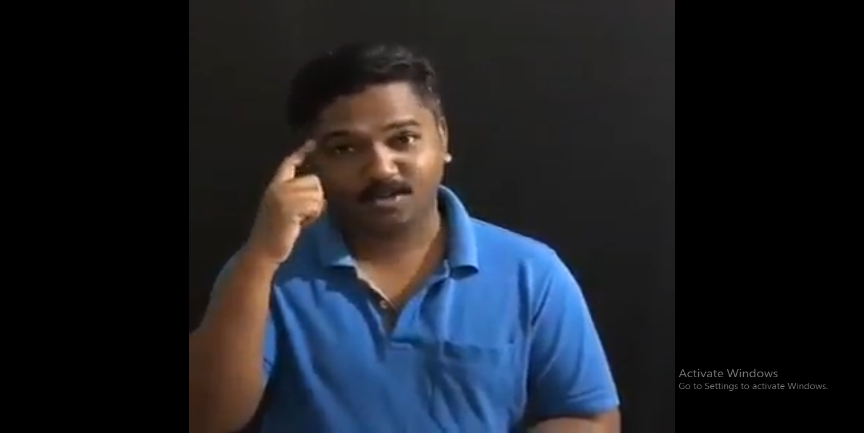}
     \end{subfigure}
     \begin{subfigure}[b]{0.3\textwidth}
         \centering
         \includegraphics[height= 3cm, width=\textwidth]{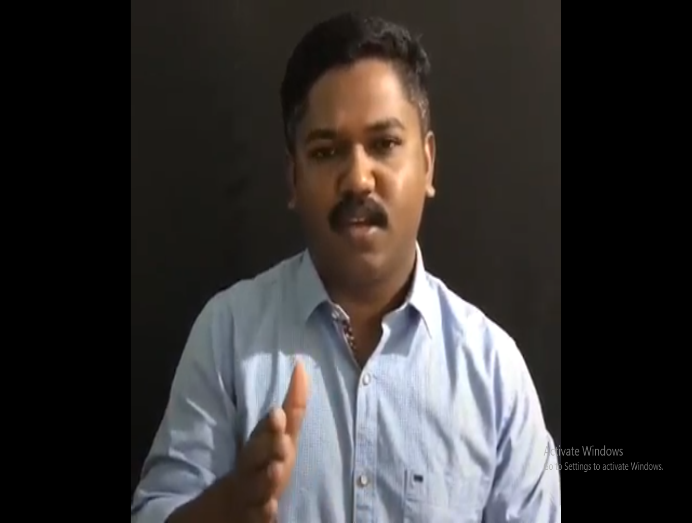}
     \end{subfigure}
        \caption{Screenshots of Malayalam video clips with negative sentiment}
        \label{fig:neg-mal}
\end{figure}

\begin{figure}
     \centering
     \begin{subfigure}[b]{0.3\textwidth}
         \centering
         \includegraphics[height= 3cm, width=\textwidth]{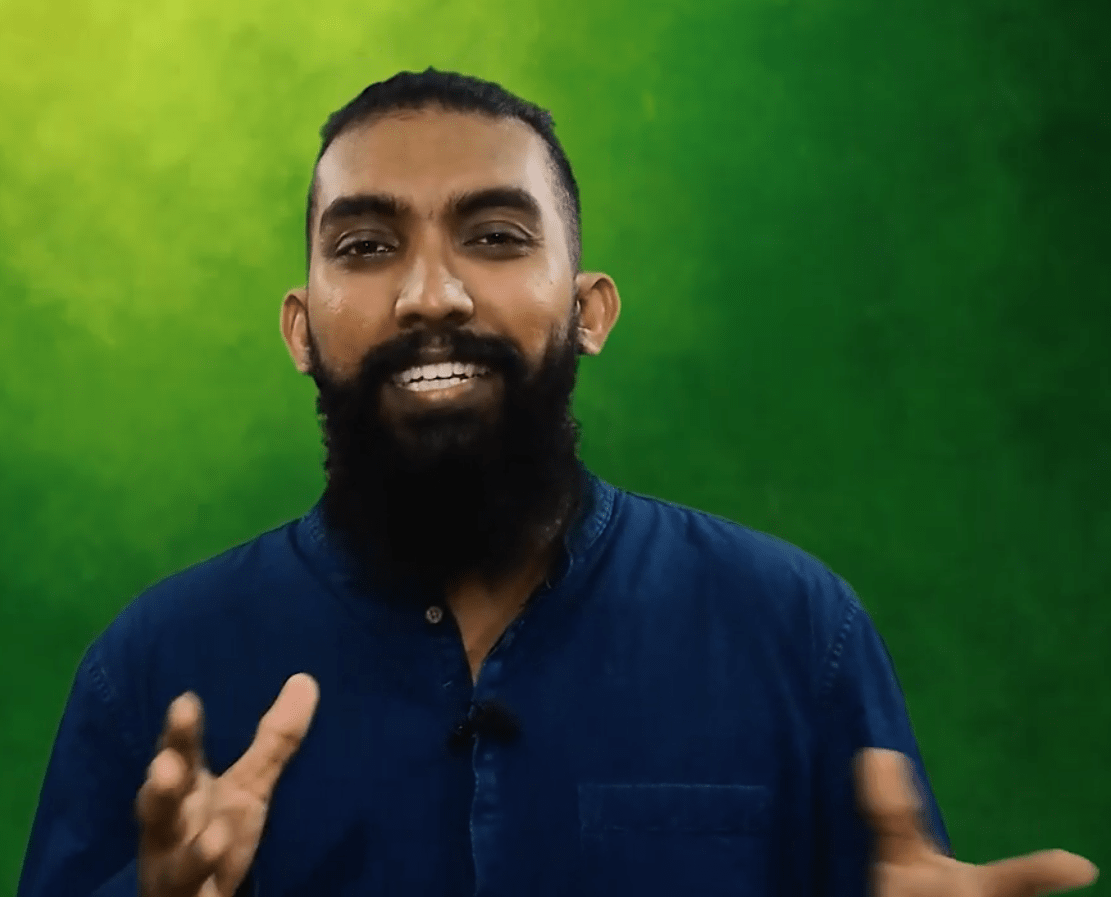}
     \end{subfigure}
     \begin{subfigure}[b]{0.3\textwidth}
         \centering
         \includegraphics[height= 3cm, width=\textwidth]{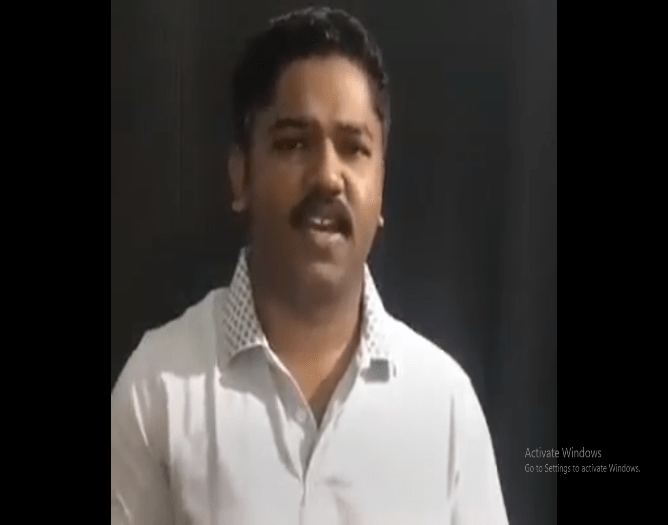}
     \end{subfigure}
        \caption{Screenshots of Malayalam video clips with highly negative sentiment}
        \label{fig:high-neg-mal}
\end{figure}

\begin{figure}
     \centering
     \begin{subfigure}[b]{0.3\textwidth}
         \centering
         \includegraphics[height= 3cm, width=\textwidth]{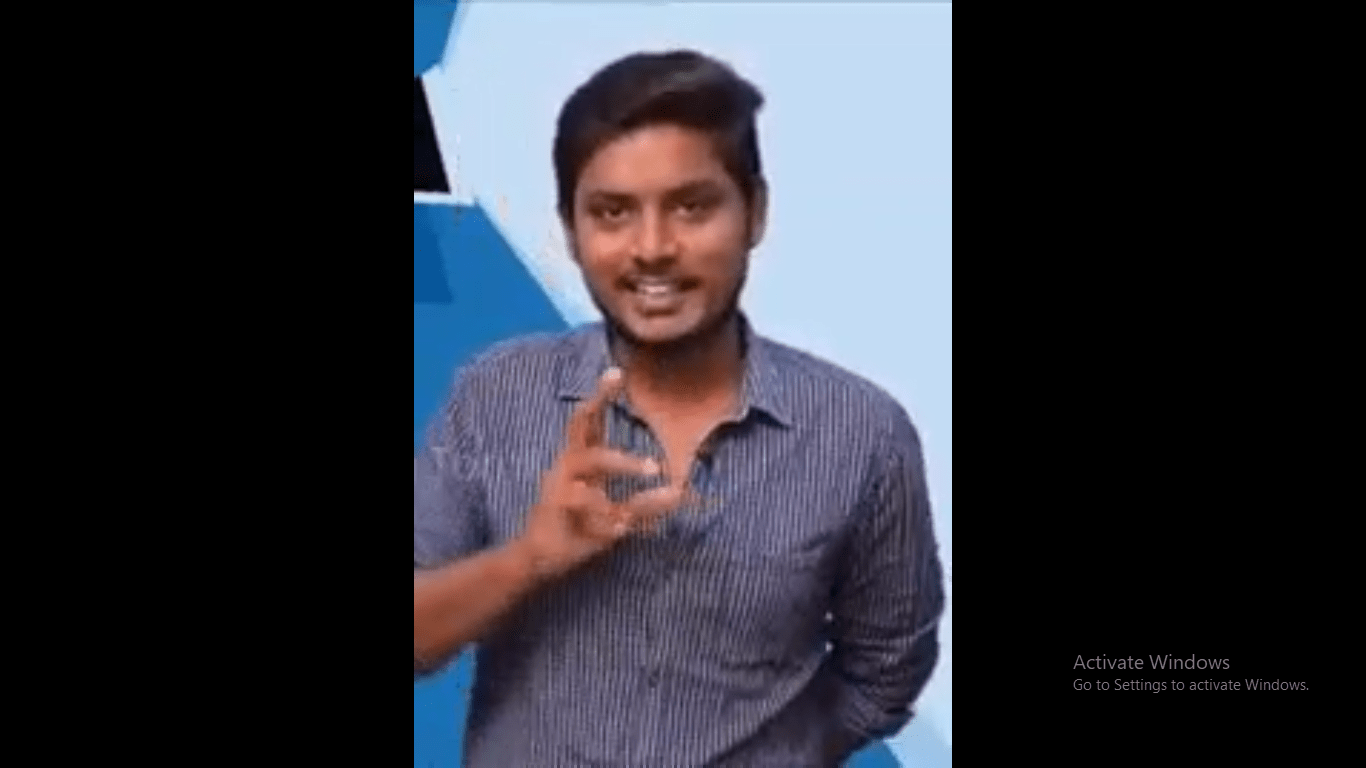}
     \end{subfigure}
     \begin{subfigure}[b]{0.3\textwidth}
         \centering
         \includegraphics[height= 3cm, width=\textwidth]{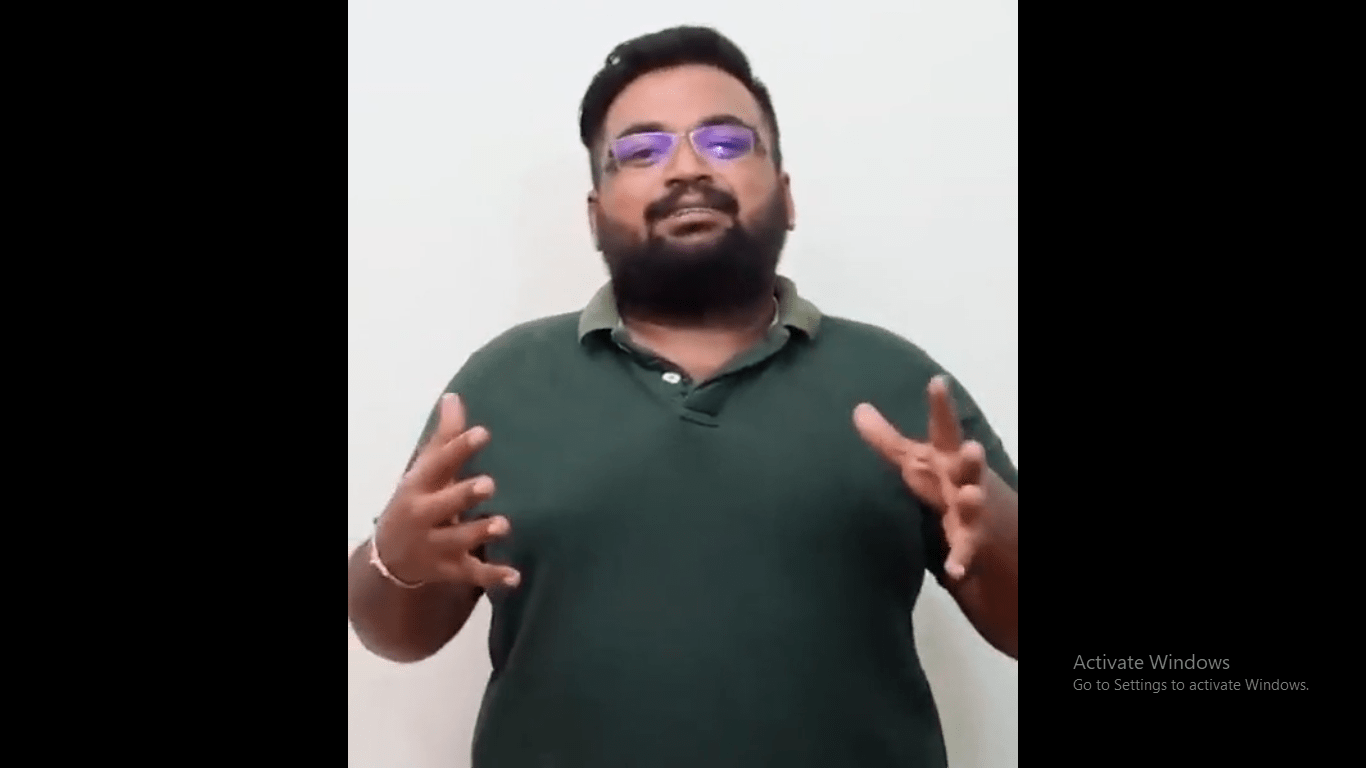}
     \end{subfigure}
        \caption{Screenshots of Tamil video clips with highly positive sentiment}
        \label{fig:high-pos-tam}
\end{figure}

\begin{figure}
     \centering
     \begin{subfigure}[b]{0.3\textwidth}
         \centering
         \includegraphics[height= 3cm, width=\textwidth]{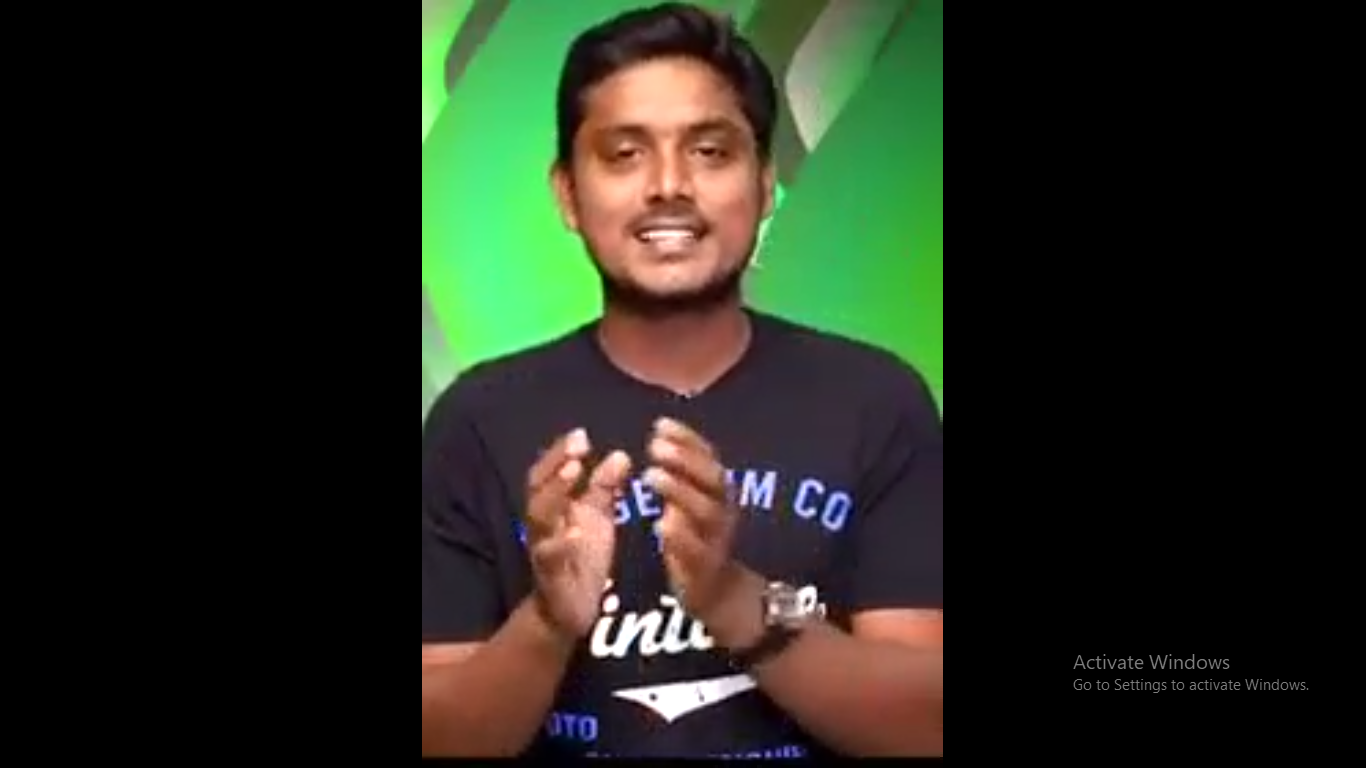}
     \end{subfigure}
     \begin{subfigure}[b]{0.3\textwidth}
         \centering
         \includegraphics[height= 3cm, width=\textwidth]{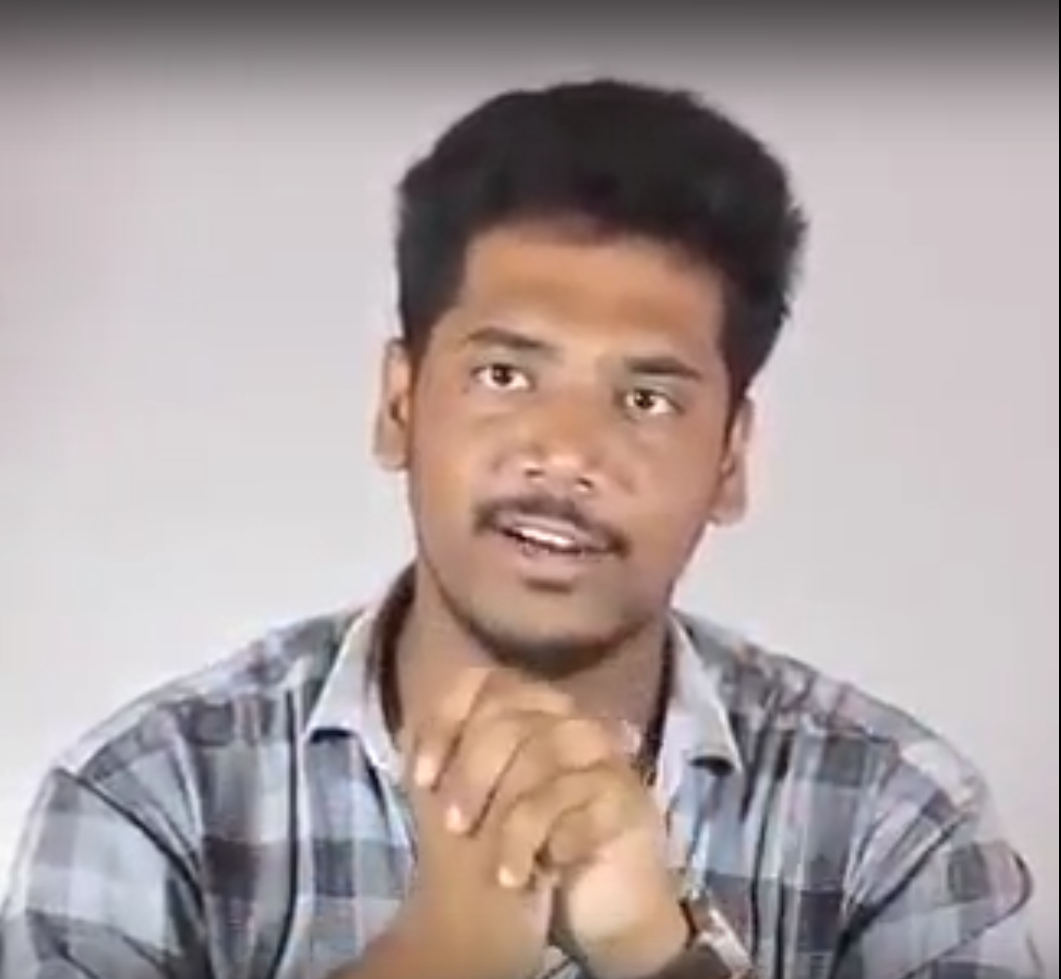}
     \end{subfigure}
        \caption{Screenshots of Tamil video clips with positive sentiment}
        \label{fig:pos-tam}
\end{figure}

\begin{figure}
     \centering
     \begin{subfigure}[b]{0.3\textwidth}
         \centering
         \includegraphics[height= 3cm, width=\textwidth]{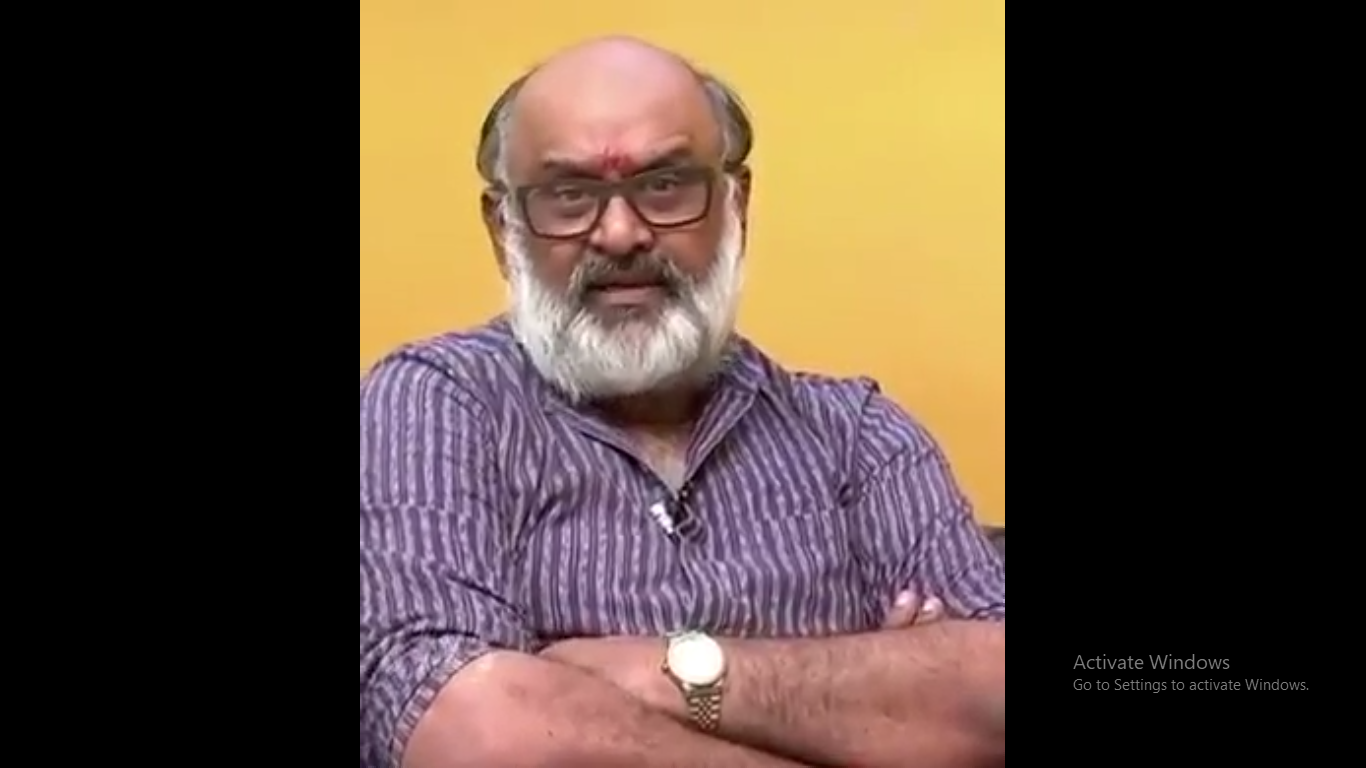}
     \end{subfigure}
     \begin{subfigure}[b]{0.3\textwidth}
         \centering
         \includegraphics[height= 3cm, width=\textwidth]{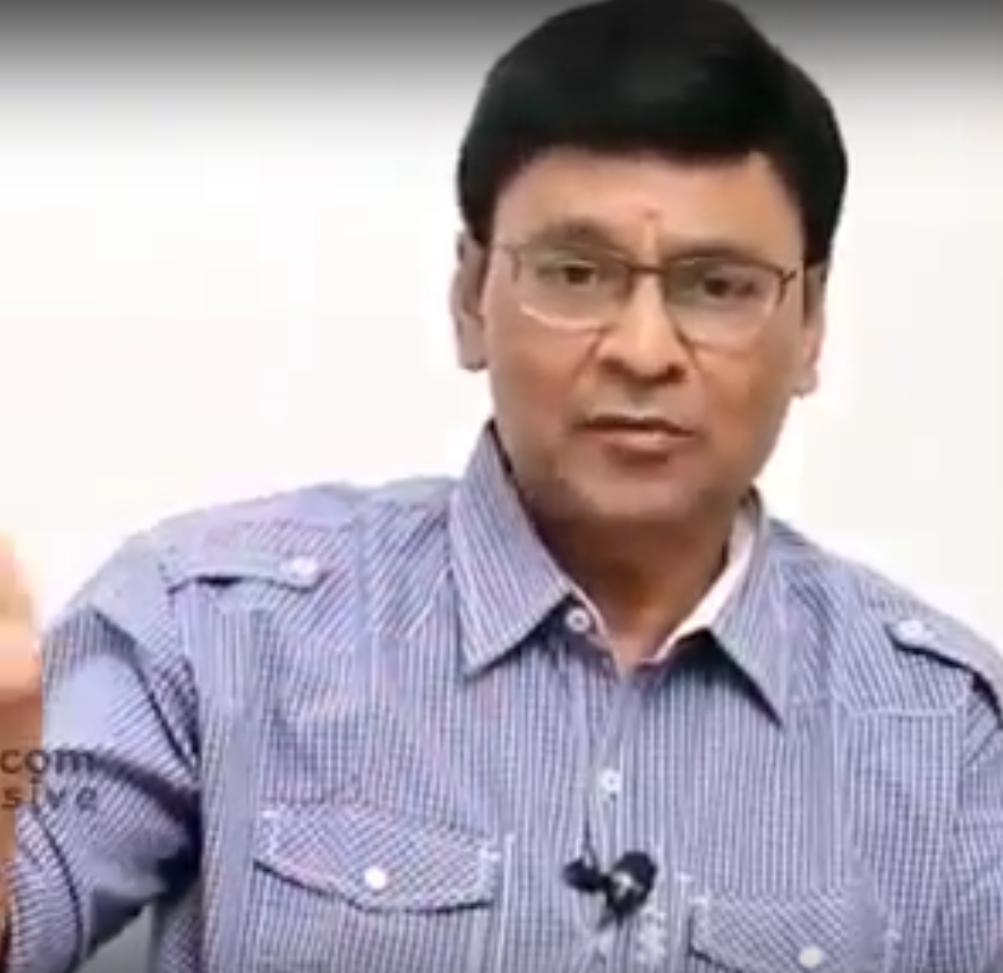}
     \end{subfigure}
        \caption{Screenshots of Tamil video clips with neutral sentiment}
        \label{fig:neu-tam}
\end{figure}

\begin{figure}
     \centering
     \begin{subfigure}[b]{0.3\textwidth}
         \centering
         \includegraphics[height= 3cm, width=\textwidth]{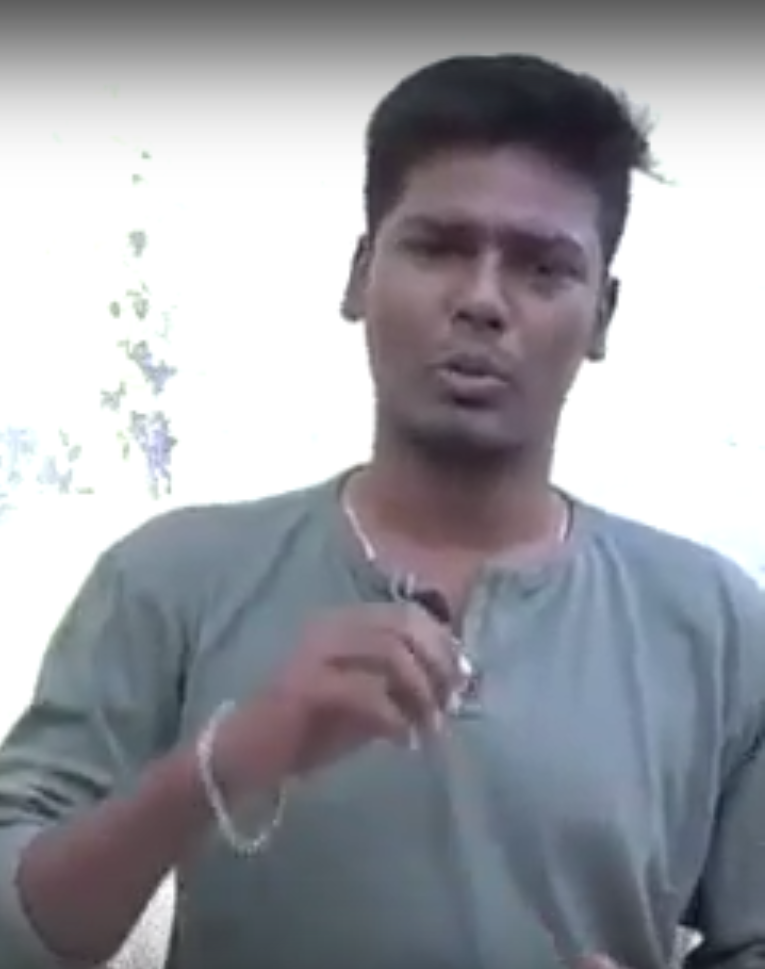}
     \end{subfigure}
     \begin{subfigure}[b]{0.3\textwidth}
         \centering
         \includegraphics[height= 3cm, width=\textwidth]{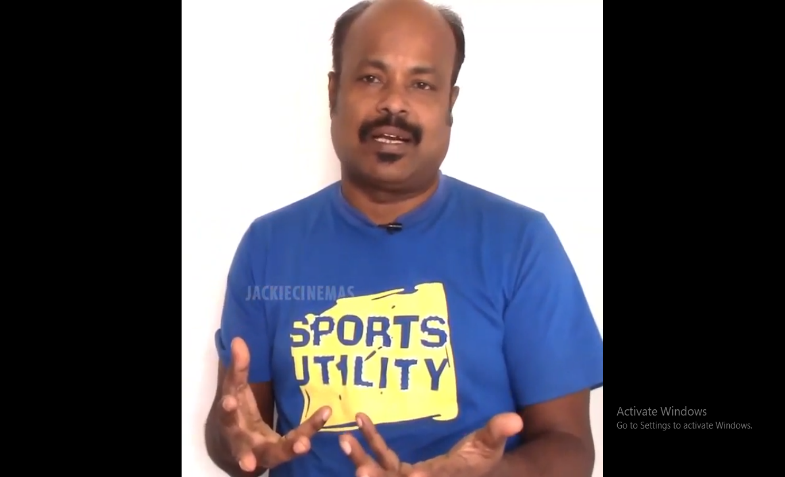}
     \end{subfigure}
        \caption{Screenshots of Tamil video clips with negative sentiment}
        \label{fig:neg-tam}
\end{figure}

\begin{figure}
     \centering
     \begin{subfigure}[b]{0.3\textwidth}
         \centering
         \includegraphics[height= 3cm, width=\textwidth]{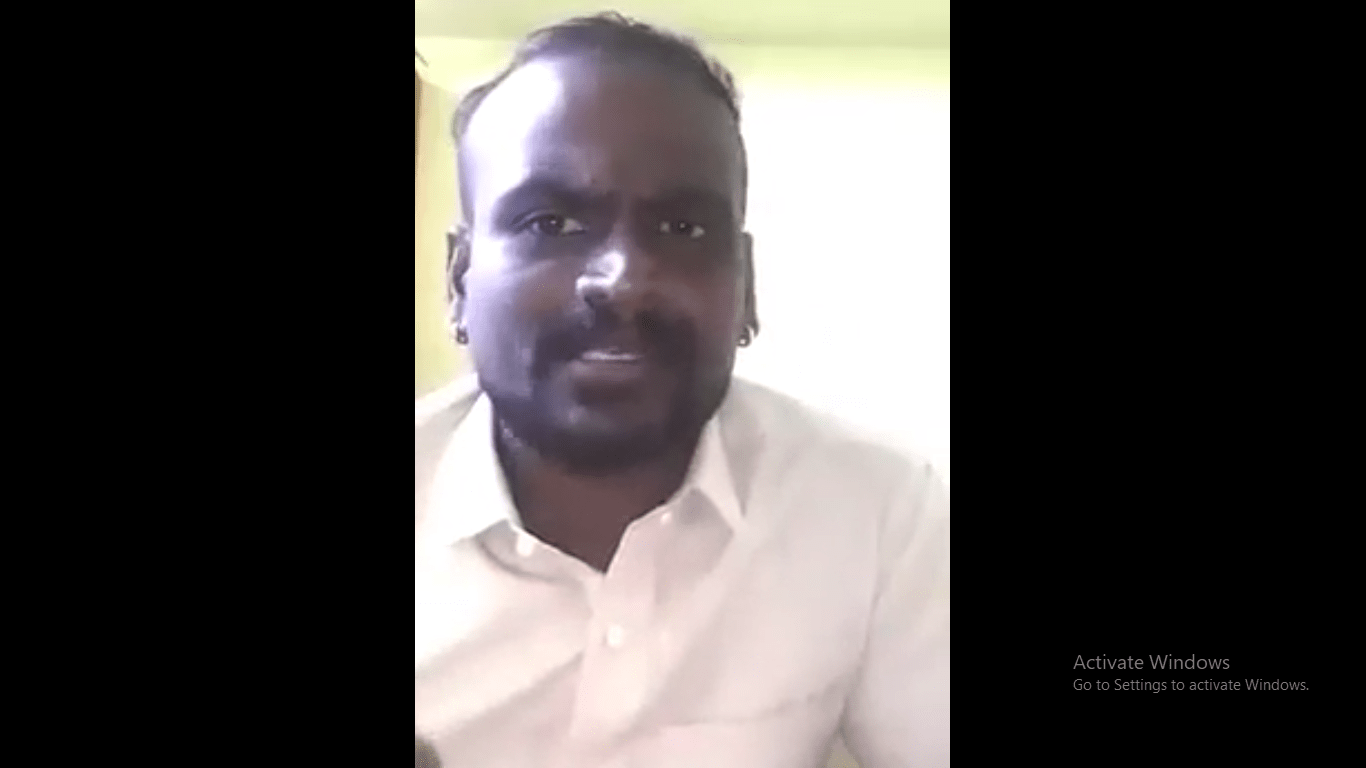}
     \end{subfigure}
     \begin{subfigure}[b]{0.3\textwidth}
         \centering
         \includegraphics[height= 3cm, width=\textwidth]{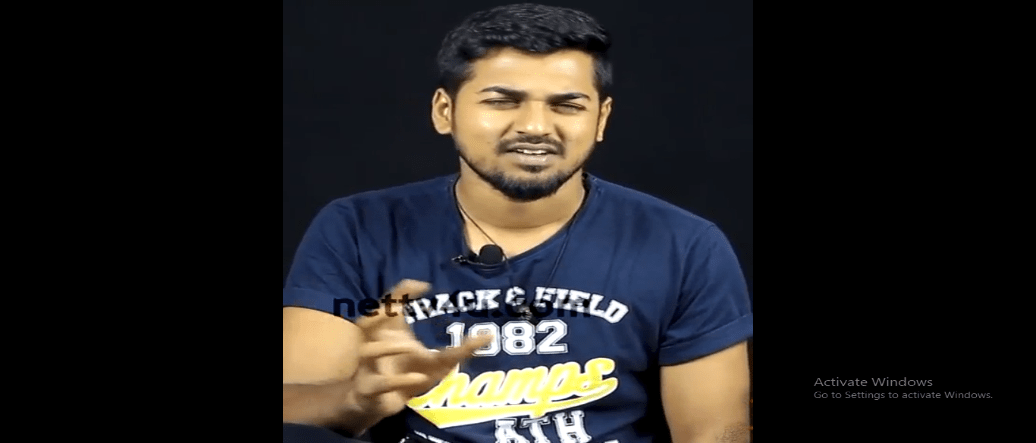}
     \end{subfigure}
        \caption{Screenshots of Tamil video clips with highly negative sentiment}
        \label{fig:high-neg-tam}
\end{figure}

Humans use facial expressions as a primary mode of communicating their emotions and sentiments. In addition to that, people use gestures to support their way of expressing feelings. Therefore, one should observe the video clips to understand the exact sentiments expressed by the reviewer. Generally, people use facial expressions to magnify their highly positive or highly negative sentiments with some gestures. It helped us to identify these sentiments quickly by taking the vocal modulation and words into consideration. We focused on the eyebrows and pupils of the eyes to pin down the sentiment. The movement and the size of the pupils, and the shape of the eyebrows helped to recognise highly positive and highly negative sentiments. Besides, the wider hand movements of the reviewer also aided in understanding the highly positive and negative sentiments. In addition to that, the amplitude of the vocal modulations was also considered. Even though the facial expressions and gestures are minimal in other cases, it was adequate to distinguish positive and negative sentiments from neutral sentiment. The connection between the facial expression, gestures, voice modulation and the words were examined for the annotation. In most of the videos, the hand gestures were not visible clearly since the video clips were cropped to remove unwanted portions.  Figures \ref{fig:high-pos-mal}, \ref{fig:pos-mal}, \ref{fig:neu-mal}, \ref{fig:neg-mal}, and \ref{fig:high-neg-mal}, show screenshots of highly positive, positive, neutral, negative, and highly negative videos in Malayalam. Similarly, \ref{fig:high-pos-tam}, \ref{fig:pos-tam}, \ref{fig:neu-tam}, \ref{fig:neg-tam} and \ref{fig:high-neg-tam} show screenshots of highly positive, positive, neutral, negative, and highly negative videos in Tamil.

\subsection{Words and sentences}
The sentiment of a sentence is greatly affected by the words used in it. Here, we initially analysed the words used in videos to understand the sentiments. Words such as `good', `bad', `very nice', `very bad', `one time watchable', `the film got awards', `a family-watchable movie', and `a super hit movie'  helped to identify the sentiments directly. Apart from that, the sentences/phrases such as “It was irritating” and “if you have a problem with someone, force him to watch this movie”  also helped to recognise the sentiments. Below are examples of highly positive, positive, negative, and highly negative words in Malayalam and Tamil data. 

\begin{itemize}
    \item Highly positive words/phrases in Malayalam
    
    Valare manoharamayi (very beautifully), anchil anch rating (five out of five ratings), kothippikkum (will be coveted)
    
    \item Positive words/phrases in Malayalam
    
    ishtappedum (would like), rasamulla (interesting), gambheeram (awsome)
    
    \item Negative words/phrases in Malayalam
    
    shokam (grief), veruppikkunna (disgusting), puthumayilla (nothing new)
    
    \item Highly negative words/phrases in Malayalam
    
    valare mosham (very bad), valareyadhikam nirasha (very disappointed), ozhivakkenda cinema (avoidable cinema)
\end{itemize}

\subsection{Direct rating}
Some reviewers give ratings to movies on a scale of five. This aspect is considered for annotation if the video contains statements related to the rating. Table \ref{tab:rating-annotation} shows the scheme followed for labelling the videos in such situations. Besides, definite reviews from the reviewers like "this movie is good" were also considered.

\begin{table}[]
    \centering
    \begin{tabular}{lc}
    \hline 
    Rating     &  Label \\ \hline
    $>=4$     & Highly Positive \\
    $>= 2.5$ and $<4$ & Positive \\
    $=2.5$ & Neutral \\
    $<2.5$ and $>1$ & Negative \\
    $<=1$ & Highly Negative \\
    \hline
    \end{tabular}
    \caption{Annotation scheme used for labelling the videos based on the ratings given by the reviewer.}
    \label{tab:rating-annotation}
\end{table}

\subsection{Attention}
The word or phrase on which the reviewer gives attention plays a vital role in determining the sentiment of a movie. A reviewer may give a positive verdict about a movie despite giving a few negative opinions. In such cases, the annotator should listen to the words to which the reviewer pays more attention. In some cases, instead of giving a direct opinion, reviewers may talk about the awards and recognition the movie has received. The annotator has to decide the sentiment from the words. We labelled such reviews as highly positive because of the appraisal given by the reviewer. Therefore, such appraisals were taken into account for annotating a video as a positive one or a highly positive one. If the reviewer makes comments similar to " It is impossible to give any rating for this movie", we annotate those reviews as negative. If the opinion of the reviewer is "It is a good film, family-oriented subject, but the scenes are predictable, and these type of stories are repeated; but old people will love it, you can watch it one time", we labelled them as neutral. Similar sentence structures were analysed for annotating the videos into any one of the five categories. 

\subsection{Sarcasm in the speech}
Sarcasm is a linguistic usage that has to be addressed carefully in applications like sentiment analysis. Generally, in reviews, people use sarcastic statements to express their negative sentiments. Sarcasm is negative sentiment in disguise. An example of such a statement is given below,

"Tears came out of my eyes by seeing such action scenes, the hero is fighting with 100 people, and the enemies are flying. Even Superman can not fight like this." 

This sentence seems to be positive but conveys a negative sentiment. Sarcasm has to be identified from the tone of the speech and the facial expression. 

\subsection{Usage of words in the speech}
Ambiguity in the senses of words is a universal problem in language. In regional languages such as Malayalam and Tamil, people use words with opposite meanings in different contexts. Dialects in these languages are one reason for such ambiguity. Apart from dialects, words used among a particular age group or locality also bring about such ambiguity. `Bhayankaram' (fearful) and `poli' (demolishing) are examples of such usage. 

\begin{table}[]
    \centering
    \begin{tabular}{lcc}
    \hline 
    Language     &  Malayalam & Tamil \\ \hline
    Number of tokens                          & 10332                  & 13066           \\
    Vocabulary size                           & 3946                   & 4445            \\
    Number of words occurs only once          & 2667                   & 3059            \\
    Number of words appear more than 50 times & 13                     & 22              \\
    Number of words appear more than 30 times & 21                     & 19              \\
    Number of words appear more than 20 times & 15                     & 38              \\
    Number of words appear more than 10 times & 98                     & 105             \\
    Age group of the speakers                & 18 to 45               & 18 to 58        \\
    Number of distinct speakers               & 20                     & 26              \\
    Male speakers                             & 19                     & 17             \\
    Female speakers                           &  1                     & 4              \\
    Video length                              & \multicolumn{2}{c}{1 to 3 minutes}       \\
    Frame width of the video                  & \multicolumn{2}{c}{342 to1280 pixels}    \\
    Frame height of the video                 & \multicolumn{2}{c}{360 to 720 pixels}    \\
    Data rate                                 & \multicolumn{2}{c}{408kps to 1152kps}    \\
    Bit rate                                  & \multicolumn{2}{c}{539 kbps to 1280kbps} \\
    Size of the videos                        & \multicolumn{2}{c}{3 MB to 55 MB}        \\
    \hline
    \end{tabular}
    \caption{Statistics of Malayalam and Tamil data across different classes.}
    \label{tab:data-distribution-2}
\end{table}

\begin{figure}
\centering
\includegraphics[scale=0.7]{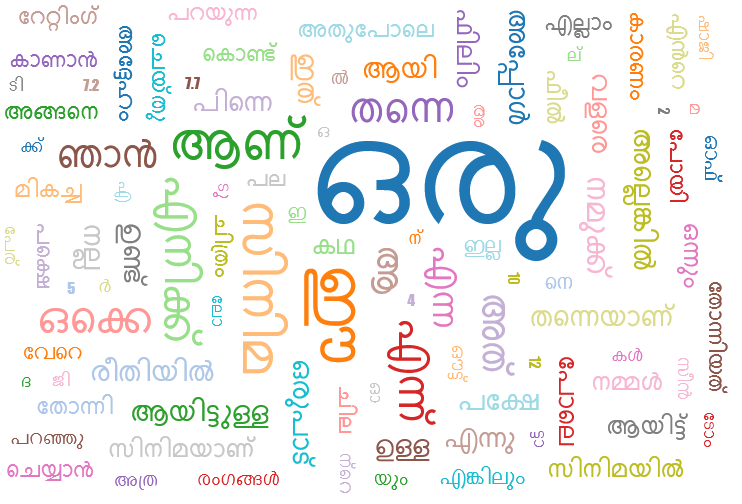}
\caption{Wordcloud for  Malayalam corpus}
\label{fig:mal-wc}
\end{figure}

\begin{figure}
\centering
\includegraphics[scale=0.7]{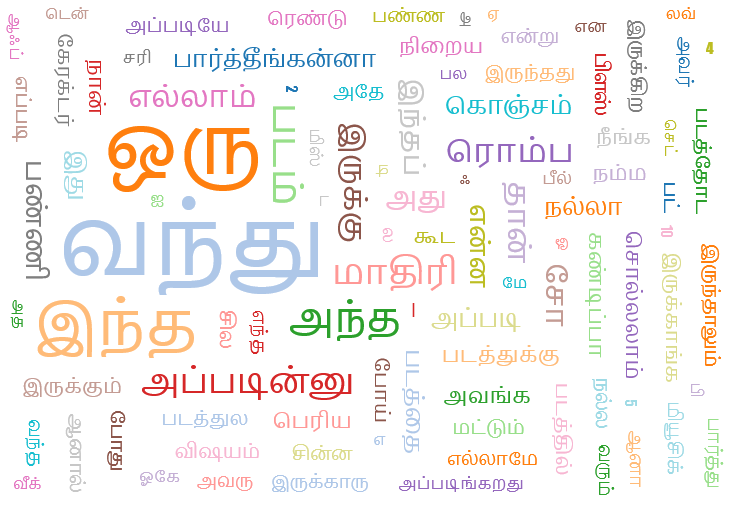}
\caption{Wordcloud for Tamil corpus}
\label{fig:tam-wc}
\end{figure}

\subsection{Inter-Annotator Agreement (IAA)}
The inter-annotator agreement is a measure of agreement between annotators while annotating the data. It also tells the clarity in the guidelines used for the annotation. The annotators tend to disagree the annotation guidelines are not well defined, and it affects the decision to find the proper annotation for data. In addition to that, it also tells how reliable an annotation is. Here Fleiss’s Kappa score is used to compute the IAA since there are three annotators.

\begin{equation}
    \kappa  = \frac{{P\left( A \right) - P\left( E \right)}}{{1 - P\left( E \right)}}
\end{equation}

Where $P\left(A\right)$ is the fraction of times all the three annotators agreed upon the same score, and $P\left(E\right)$ is the probability of the expected agreement. We followed the definition given by \citewpar{landis1977measurement} to interpret the Kappa scores, which is given in Table \ref{tab:kappa}. 

\begin{table}[]
    \centering
    \begin{tabular}{|l|r|}
    \hline 
    Kappa value     &  Interpretation \\ \hline
    $<0.00$ & Poor agreement  \\
    $0.00-0.20 $ & Slight agreement \\
    $0.21-0.40$ & Fair agreement \\
    $0.41-0.60$ & Moderate agreement \\
    $0.61-0.80$ & Substantial agreement \\
    $0.81-1.00 $ & Almost perfect or perfect agreement \\
    \hline
    \end{tabular}
    \caption{Interpretation of the agreement based on Kappa scores.}
    \label{tab:kappa}
\end{table}

The number of times the annotators were ambiguous in identifying the proper label was less. Mainly, the ambiguity came in Positive-Highly Positive, Positive-Neutral, and Negative-Highly Negative pairs. In these cases, the labels with maximum votes were taken as the label of the data. Table \ref{tab:kappa-score} gives the Kappa score for IAA in Malayalam and Tamil data. The Kappa score for Malayalam and Tamil are 0.7307 and 0.7496, respectively. Therefore, according to the agreement interpretation rules, we can conclude that there is a substantial agreement between the annotators in labelling the video clips into five different sentiments.  

\begin{table}[]
    \centering
    \begin{tabular}{|l|r|}
    \hline 
    Language     &  Kappa score \\ \hline
    Malayalam  & 0.7307  \\
    Tamil & 0.7496 \\
    \hline
    \end{tabular}
    \caption{Kappa score for Malayalam and Tamil data.}
    \label{tab:kappa-score}
\end{table}

\section{Corpus Statistics} \label{Corpusstat}
In Table \ref{tab:data-distribution-2}, we provide an overview of specific statistics of our dataset such as number of tokens, vocabulary size, number of words appear at different times, details about the speakers, frame width, height of video, data rate, bit rate, and size of the video clips. From the Table \ref{tab:data-distribution-2}, we can see that our dataset contains equal number of speakers in case of gender. Our video length is limited to 1 to 3 minutes because we post-processed to get the short clip of the whole as discussed in the previous sections.  Figure \ref{fig:mal-wc} show the most frequent word in the corpus of our dataset in Malayalam languages as `oru' meaning `one or a' in English. Since most of the short speakers in the short clip talk about about one movie or starts a sentence with.  In case of Tamil,  multiple words comes a most frequent words including `oru, vanthu, intha' meaning `one or a, came, this' which shows diversity of word usage in Tamil. 
\begin{table}[]
    \centering
    \begin{tabular}{|l|r|r|}
    \hline 
    Class label     &  Malayalam & Tamil \\ \hline
    Highly Positive & 9 (12.85\%) & 8 (12.50\%) \\
    Positive & 39 (55.71\%) & 38 (59.37\%)\\
    Neutral & 8 (11.42\%) & 8 (12.50\%) \\
    Negative & 12 (17.14\%) & 5 (7.81\%)\\
    Highly Negative & 2 (2.85\%) & 5 (7.81\%)\\
    Total & 70 & 64 \\
    \hline
    \end{tabular}
    \caption{Distribution of Malayalam and Tamil data across different classes.}
    \label{tab:data-distribution-1}
\end{table}

\begin{table}[]
    \centering
    \begin{tabular}{|l|r|r|}
    \hline 
    Sentiment     &  Malayalam & Tamil \\ \hline
    Highly Positive & 82 & 87 \\
    Positive & 193 & 182 \\
    Negative & 99 & 89 \\
    Highly Negative & 27 & 15 \\
    \hline
    \end{tabular}
    \caption{Statistics of the Malayalam and Tamil words/phrases with different sentiments.}
    \label{tab:data-sentiment-stat}
\end{table}

In Table \ref{tab:data-distribution-1}, we provide class-wise distribution of the sentiment annotations. Our dataset is more skewed towards positive nearly more than half are positive out of five classes. This might be due to fact the reviewer take movie to review based on high ratings or highest suggestion so the reviewers are biased towards selecting popular movie to review so that they can get subscriptions. The extreme case of highly negative is very low (2.85\%) for Malayalam. However, for Tamil we have same as of negative. Our dataset contains similar percentage of distribution of classes for both languages.  Percentage wise distribution are shown in the the Table \ref{tab:data-distribution-1}. The number of words or phrases with different sentiment senses is given in Table \ref{tab:data-sentiment-stat}. Positive and negative words account for 4.89\% and 2.51\% in Malayalam, and 4.09\% and 2\%. Highly positive and highly negative sentiments were identified using either words or noun phrases. In this scenario, a noun phrase consists of an adjective and a noun.

As the interest towards automatic identification sentiment and opinion in video is increasing, the videos in the presented dataset are intended to train and benchmark techniques for multimodal sentiment analysis in under-resourced Tamil and Malayalam languages.

\begin{table}[]
\begin{tabular}{ll}

\hline
File name    & YouTube link     \\ \hline
MAL\_MSA\_01 & \url{https://www.youtube.com/watch?v=OEdXwkY9Hy8}  \\
MAL\_MSA\_02 & \url{https://www.youtube.com/watch?v=HBN1TlHrNFM}  \\
MAL\_MSA\_03 & \url{https://www.youtube.com/watch?v=BgWuMofZavg}  \\
MAL\_MSA\_04 & \url{https://www.youtube.com/watch?v=LpX\_zaDXyK8}    \\
MAL\_MSA\_05 & \url{https://www.youtube.com/watch?v=KW-pS5uZOXI}  \\
MAL\_MSA\_06 & \url{https://www.youtube.com/watch?v=3LhK-IrOVIs}  \\
MAL\_MSA\_07 & \url{https://www.youtube.com/watch?v=8NLvtcJ-v9k}  \\
MAL\_MSA\_08 & \url{https://www.youtube.com/watch?v=m1FMnc7BAw4\&t=127s}        \\
MAL\_MSA\_09 & \url{https://www.youtube.com/watch?v=RpeD\_y0Z8dA\&t=75s}        \\
MAL\_MSA\_10 & \url{https://www.youtube.com/watch?v=ylKc8Nn7DlY\&t=63s}         \\
MAL\_MSA\_11 & \url{https://www.youtube.com/watch?v=II2zwF0m2c0\&t=76s}         \\
MAL\_MSA\_12 & \url{https://www.youtube.com/watch?v=6qCWxVpU6sk\&t=13s}         \\
MAL\_MSA\_13 & \url{https://www.youtube.com/watch?v=keyu\_Dbyj50}    \\
MAL\_MSA\_14 & \url{https://www.youtube.com/watch?v=-zcjhnptFzY}  \\
MAL\_MSA\_15 & \url{https://www.youtube.com/watch?v=dSLAWrZ50Ws}  \\
MAL\_MSA\_16 & \url{https://www.youtube.com/watch?v=iUR206f1js8\&t=41s}         \\
MAL\_MSA\_17 & \url{https://www.youtube.com/watch?v=dN6Nd4u2JSg\&t=30s}         \\
MAL\_MSA\_18 & \url{https://www.youtube.com/watch?v=D-Kn0k9LWaQ\&t=50s}         \\
MAL\_MSA\_19 & \url{https://www.youtube.com/watch?v=pSU\_KF0dEeQ\&t=36s}        \\
MAL\_MSA\_20 & \url{https://www.youtube.com/watch?v=4oNHvAff72g}  \\
MAL\_MSA\_21 & \url{https://www.youtube.com/watch?v=1h5m6bSXjjE}  \\
MAL\_MSA\_22 & \url{https://www.youtube.com/watch?v=zqgaz5EvZb8}\\
MAL\_MSA\_23 & \url{https://www.youtube.com/watch?v=SY4Ma2MiLbU }\\
MAL\_MSA\_24 & \url{https://www.youtube.com/watch?v=8rGTivqCl30\&t=60s}         \\
MAL\_MSA\_25 & \url{https://www.youtube.com/watch?v=sUgEIVVSgRg\&t=42s}         \\
MAL\_MSA\_26 & \url{https://www.youtube.com/watch?v=w5y\_cNKUuTo}    \\
MAL\_MSA\_27 & \url{https://www.youtube.com/watch?v=QTGwrkCPb6I\&t=37s}         \\
MAL\_MSA\_28 & \url{https://www.youtube.com/watch?v=IYvwwts6ylc}  \\
MAL\_MSA\_29 & \url{https://www.youtube.com/watch?v=Z8vN-lICQpY\&t=95s}         \\
MAL\_MSA\_30 & \url{https://www.youtube.com/watch?v=5BhbStbRbiY}  \\
MAL\_MSA\_31 & \url{https://www.youtube.com/watch?v=KxsRPYzfGL8}  \\
MAL\_MSA\_32 & \url{https://www.youtube.com/watch?v=PB1rYVRPpIc}  \\
MAL\_MSA\_33 & \url{https://www.youtube.com/watch?v=kZDvEB5vbws\&t=1s}          \\
MAL\_MSA\_34 & \url{https://www.youtube.com/watch?v=-TYug7DWrUU}  \\
MAL\_MSA\_35 & \url{https://www.youtube.com/watch?v=QldKGlhssNg}  \\
MAL\_MSA\_36 & \url{https://www.youtube.com/watch?v=OUqLkIyBCAc}  \\
MAL\_MSA\_37 & \url{https://www.youtube.com/watch?v=HMmf-qz5E3A}  \\
MAL\_MSA\_38 & \url{https://www.youtube.com/watch?v=8xsQlhPTkzg\&t=17s}         \\
MAL\_MSA\_39 & \url{https://www.youtube.com/watch?v=oyiwDzn3wXc\&t=42s}         \\
MAL\_MSA\_40 & \url{https://www.youtube.com/watch?v=51\_8LymZjYo\&t=15s}        \\
MAL\_MSA\_41 & \url{https://www.youtube.com/watch?v=pvSG2Ys\_b0U\&t=5s}         \\
MAL\_MSA\_42 & \url{https://www.youtube.com/watch?v=I-\_S5wvkPlU}    \\
MAL\_MSA\_43 & \url{https://www.youtube.com/watch?v=C2cbwzsf1Iw}  \\
MAL\_MSA\_44 & \url{https://www.youtube.com/watch?v=NMwEnE5QkHI\&t=49s}         \\
MAL\_MSA\_45 & \url{https://www.youtube.com/watch?v=\_lmUmLLkYsM}    \\
MAL\_MSA\_46 & \url{https://www.youtube.com/watch?v=L-r10YFSsQU}  \\
MAL\_MSA\_47 & \url{https://www.youtube.com/watch?v=cjGais3y8Ts}  \\
MAL\_MSA\_48 & \url{https://www.youtube.com/watch?v=LwPcT\_UGSwE}    \\
MAL\_MSA\_49 & \url{https://www.youtube.com/watch?v=54G\_1ErD6Vk}    \\
MAL\_MSA\_50 & \url{https://www.youtube.com/watch?v=K8XOQT4d1ME\&t=55s}         \\
MAL\_MSA\_51 & \url{https://www.youtube.com/watch?v=v36d1Q9eUko}  \\
MAL\_MSA\_52 & \url{https://www.youtube.com/watch?v=hLmidefhmew}  \\
MAL\_MSA\_53 & \url{https://www.youtube.com/watch?v=MJBXH4PxZYM}  \\
MAL\_MSA\_54 & \url{https://www.youtube.com/watch?v=v2ckh3CkYJg}  \\
MAL\_MSA\_55 & \url{https://www.youtube.com/watch?v=TwCPlpP7CBQ\&t=28s}         \\
MAL\_MSA\_56 & \url{https://www.youtube.com/watch?v=QiM5UPUgk2M\&t=4s}          \\
MAL\_MSA\_57 & \url{https://www.youtube.com/watch?v=e4OJlY-rKDo\&t=12s}         \\
MAL\_MSA\_58 & \url{https://www.youtube.com/watch?v=n7brqi13nTY}  \\
MAL\_MSA\_59 & \url{https://www.youtube.com/watch?v=lVO\_2Huy\_Xs}\\
MAL\_MSA\_60 & \url{https://www.youtube.com/watch?v=ltL2vxMeX20}           \\
MAL\_MSA\_61 & \url{https://www.youtube.com/watch?v=fpUH2r85dWo}  \\
MAL\_MSA\_62 & \url{https://www.youtube.com/watch?v=dciM83P43SE\&t=26s}         \\
MAL\_MSA\_63 & \url{https://www.youtube.com/watch?v=ZVfSoYxf41Q\&t=45s}         \\
MAL\_MSA\_64 & \url{https://www.youtube.com/watch?v=vvVA9rfr920\&t=37s}         \\
MAL\_MSA\_65 & \url{https://www.youtube.com/watch?v=72Wud10Vv2E}  \\
MAL\_MSA\_66 & \url{https://www.youtube.com/watch?v=IWnYM\_Bf9vI\&t=24s}        \\
MAL\_MSA\_67 & \url{https://www.youtube.com/watch?v=o5ndBF-yU4c\&t=117s}        \\
MAL\_MSA\_68 & \url{https://www.youtube.com/watch?v=0ZfWu0iY6LI\&t=198s}        \\
MAL\_MSA\_69 & \url{https://www.youtube.com/watch?v=UP5AMOalIq0\&t=18s}         \\
MAL\_MSA\_70 & \url{https://www.youtube.com/watch?v=jHGhkorv8eU} \\ \hline
\end{tabular}
\caption{YouTube URLs for Malayalam videos}
\label{tab:mal_yt}
\end{table}

\begin{table}[]
\begin{tabular}{ll}
\hline
File name    & YouTube link                                  \\ \hline
TAM\_MSA\_01 & \url{https://www.youtube.com/watch?v=A7hmrsDl0ho}   \\
TAM\_MSA\_02 & \url{https://www.youtube.com/watch?v=CWz6srpNvR4}   \\
TAM\_MSA\_03 & \url{https://www.youtube.com/watch?v=Y1gKRoXhhxA}   \\
TAM\_MSA\_04 & \url{https://www.youtube.com/watch?v=P\_YHnQMHj\_Q\&t=20s}        \\
TAM\_MSA\_05 & \url{https://www.youtube.com/watch?v=vkcqX7RmPZc}   \\
TAM\_MSA\_06 & \url{https://www.youtube.com/watch?v=VDmOj4vM688}   \\
TAM\_MSA\_07 & \url{https://www.youtube.com/watch?v=UDt2U2VzroI}   \\
TAM\_MSA\_08 & \url{https://www.youtube.com/watch?v=oVZO\_lrbIuI}\\
TAM\_MSA\_09 & \url{https://www.youtube.com/watch?v=T7JEKAl-iZc}    \\
TAM\_MSA\_10 & \url{https://www.youtube.com/watch?v=Eob7AKG0\_v8}   \\
TAM\_MSA\_11 & \url{https://www.youtube.com/watch?v=bvc9PuJZjCk}    \\
TAM\_MSA\_12 & \url{https://www.youtube.com/watch?v=0\_I\_P2FnBD0}    \\
TAM\_MSA\_13 & \url{https://www.youtube.com/watch?v=hIClEzwfDLM}   \\
TAM\_MSA\_14 & \url{https://www.youtube.com/watch?v=g5z7QKoQXmk}   \\
TAM\_MSA\_15 & \url{https://www.youtube.com/watch?v=QhA8Mg1PaFs}   \\
TAM\_MSA\_16 & \url{https://www.youtube.com/watch?v=RvGiRU\_DzZU}  \\
TAM\_MSA\_17 & \url{https://www.youtube.com/watch?v=MyZMaB7cw7M}   \\
TAM\_MSA\_18 & \url{https://www.youtube.com/watch?v=IkjBtk4ATb8\&t=75s}          \\
TAM\_MSA\_19 & \url{https://www.youtube.com/watch?v=qqDViMLwD4c}   \\
TAM\_MSA\_20 & \url{https://www.youtube.com/watch?v=g0pCn-xVtvw}   \\
TAM\_MSA\_21 & \url{https://www.youtube.com/watch?v=NYlyhtYoxXc}   \\
TAM\_MSA\_22 & \url{https://www.youtube.com/watch?v=x3RIKrr--GU}   \\
TAM\_MSA\_23 & \url{https://www.youtube.com/watch?v=PcYSjfW00Iw\&t=241s}         \\
TAM\_MSA\_24 & \url{https://www.youtube.com/watch?v=g7WVSkrCdd0\&t=54s}          \\
TAM\_MSA\_25 & \url{https://www.youtube.com/watch?v=IDnHYEuocn4\&t=32s}          \\
TAM\_MSA\_26 & \url{https://www.youtube.com/watch?v=bqK2w2QXmlA\&t=34s}          \\
TAM\_MSA\_27 & \url{https://www.youtube.com/watch?v=FOXZIGTdZpM}   \\
TAM\_MSA\_28 & \url{https://www.youtube.com/watch?v=NzsTFyYTO4g\&t=15s}          \\
TAM\_MSA\_29 & \url{https://www.youtube.com/watch?v=cAyJAbD4nuA\&t=6s}           \\
TAM\_MSA\_30 & \url{https://www.youtube.com/watch?v=5gGbW8Fr-wU\&t=47s}          \\
TAM\_MSA\_31 & \url{https://www.youtube.com/watch?v=W3ezQDkCjbw\&t=28s}          \\
TAM\_MSA\_32 & \url{https://www.youtube.com/watch?v=zgouiayTHhk}   \\
TAM\_MSA\_33 & \url{https://www.youtube.com/watch?v=GVx-xBpylU0\&t=86s}          \\
TAM\_MSA\_34 & \url{https://www.youtube.com/watch?v=hp0FS6sJql0\&t=14s}          \\
TAM\_MSA\_35 & \url{https://www.youtube.com/watch?v=qbXcQxKTuYM\&t=24s}          \\
TAM\_MSA\_36 & \url{https://www.youtube.com/watch?v=0daf2xJ0Vuw\&t=38s}          \\
TAM\_MSA\_37 & \url{https://www.youtube.com/watch?v=v2KI51rCoAA\&t=124s}         \\
TAM\_MSA\_38 & \url{https://www.youtube.com/watch?v=snwGKQiz40E\&t=53s}          \\
TAM\_MSA\_39 & \url{https://www.youtube.com/watch?v=ehJaI1O4yZY\&t=48s}          \\
TAM\_MSA\_40 & \url{https://www.youtube.com/watch?v=lzhT1cTg2VM\&t=20s}          \\
TAM\_MSA\_41 & \url{https://www.youtube.com/watch?v=00abZA6vN90\&t=32s}          \\
TAM\_MSA\_42 & \url{https://www.youtube.com/watch?v=HJMWQAK\_XlI\&t=142s}        \\
TAM\_MSA\_43 & \url{https://www.youtube.com/watch?v=uJDY\_hcv0g4\&t=74s}         \\
TAM\_MSA\_44 & \url{https://www.youtube.com/watch?v=ZUjRXXE\_Xmc\&t=113s}        \\
TAM\_MSA\_45 & \url{https://www.youtube.com/watch?v=lCU7O0qNPHU\&t=102s}         \\
TAM\_MSA\_46 & \url{https://www.youtube.com/watch?v=JP7s0ekmORM}   \\
TAM\_MSA\_47 & \url{https://www.youtube.com/watch?v=4MnlPele8c8\&t=39s}          \\
TAM\_MSA\_48 & \url{https://www.youtube.com/watch?v=wyFBw2X2vjc\&t=3s}           \\
TAM\_MSA\_49 & \url{https://www.youtube.com/watch?v=up93B\_dBahQ}  \\
TAM\_MSA\_50 & \url{https://www.youtube.com/watch?v=PAaFa798fCw\&t=12s}          \\
TAM\_MSA\_51 & \url{https://www.youtube.com/watch?v=EHQvQeaC7YA}  \\
TAM\_MSA\_52 & \url{https://www.youtube.com/watch?v=XL8YC8vS8O0\&t=33s}          \\
TAM\_MSA\_53 & \url{https://www.youtube.com/watch?v=OSTYwklSwAU\&t=2s}           \\
TAM\_MSA\_54 & \url{https://www.youtube.com/watch?v=kn5SOngitXY\&t=6s}           \\
TAM\_MSA\_55 & \url{https://www.youtube.com/watch?v=emXaS1UaIYw}   \\
TAM\_MSA\_56 & \url{https://www.youtube.com/watch?v=ApC1YEKWlhk}    \\
TAM\_MSA\_57 & \url{https://www.youtube.com/watch?v=LOQ0dei1RzE}   \\
TAM\_MSA\_58 & \url{https://www.youtube.com/watch?v=idYvwQIILw0\&t=1s}           \\
TAM\_MSA\_59 & \url{https://www.youtube.com/watch?v=jxuLOStIaO0}   \\
TAM\_MSA\_60 & \url{https://www.youtube.com/watch?v=KdpdYuHFbPU}  \\
TAM\_MSA\_61 & \url{https://www.youtube.com/watch?v=Hb2lF1EynNQ}  \\
TAM\_MSA\_62 & \url{https://www.youtube.com/watch?v=fT6fZzva2WY}           \\
TAM\_MSA\_63 & \url{https://www.youtube.com/watch?v=MiaCCNudy1g\&t=23s}          \\
TAM\_MSA\_64 & \url{https://www.youtube.com/watch?v=pTx5De-dWFE\&t=79s}   \\ \hline     
\end{tabular}
\caption{YouTube URLs for Tamil videos}
\label{tab:tam_yt}
\end{table}


\section{Conclusion} \label{Conclusion}
In this paper, we presented the new dataset for multimodal sentiment analysis called DravidianMultiModality consists of 134 videos annotated by volunteer anaotators from  online speakers out of which 70 are Malayalam videos and 64 are Tamil videos. The dataset is the first multimodal sentiment analysis dataset with sentiment polarity for Dravidian languages. We believe that the presented dataset establishes a valuable contribution to the Dravidian language research community and hopes that this dataset also opens the door to more studies on creating a multimodal dataset for under-resourced languages.  In future work, we plan to expand the dataset to other Dravidian languages, and we believe this will be a stepping stone for the researchers in the multimodal domain for Dravidian languages. 
\begin{acknowledgements}
If you'd like to thank anyone, place your comments here
and remove the percent signs.
\end{acknowledgements}


\section*{Conflict of interest}
The authors declare that they have no conflict of interest.

\bibliographystyle{spbasic}      
\bibliography{ref,anthology}  


\end{document}